\documentclass[times,twocolumn,final,authoryear]{elsarticle}

\usepackage{ycviu}

\usepackage{amsmath,amssymb,amsfonts}
\usepackage{graphicx}
\usepackage{adjustbox}
\usepackage{framed,multirow}
\usepackage{subfig}
\usepackage{xcolor}
\usepackage{url}
\newcommand{\bftab}{\fontseries{b}\selectfont}
\definecolor{darkgreen}{RGB}{23,168,23}
\usepackage{dblfloatfix}
\usepackage[breaklinks=true,bookmarks=false,colorlinks]{hyperref}
\usepackage[symbol]{footmisc}
\usepackage{natbib}
\usepackage{etoolbox}
\usepackage{latexsym}
\usepackage{lineno}

\journal{Computer Vision and Image Understanding}

\makeatletter

\patchcmd{\NAT@citex}
  {\@citea\NAT@hyper@{%
     \NAT@nmfmt{\NAT@nm}%
     \hyper@natlinkbreak{\NAT@aysep\NAT@spacechar}{\@citeb\@extra@b@citeb}%
     \NAT@date}}
  {\@citea\NAT@nmfmt{\NAT@nm}%
   \NAT@aysep\NAT@spacechar\NAT@hyper@{\NAT@date}}{}{}

\patchcmd{\NAT@citex}
  {\@citea\NAT@hyper@{%
     \NAT@nmfmt{\NAT@nm}%
     \hyper@natlinkbreak{\NAT@spacechar\NAT@@open\if*#1*\else#1\NAT@spacechar\fi}%
       {\@citeb\@extra@b@citeb}%
     \NAT@date}}
  {\@citea\NAT@nmfmt{\NAT@nm}%
   \NAT@spacechar\NAT@@open\if*#1*\else#1\NAT@spacechar\fi\NAT@hyper@{\NAT@date}}
  {}{}

\makeatother

\defcitealias{companion}{GMS}

\begin{document}

\begin{frontmatter}

\title{When CNNs Meet Random RNNs: Towards Multi-Level Analysis for \\RGB-D Object and Scene Recognition}

\author[1]{Ali Caglayan\corref{cor1}} 
\cortext[cor1]{Corresponding author:}
\ead{firstname.lastname@aist.go.jp}
\author[1]{Nevrez Imamoglu}
\author[2]{Ahmet Burak Can}
\author[1]{Ryosuke Nakamura}

\address[1]{National Institute of Advanced Industrial Science and Technologhy (AIST), Tokyo, Japan}
\address[2]{Department of Computer Engineering, Hacettepe University, Ankara, Turkey}

\received{1 May 2013}
\finalform{10 May 2013}
\accepted{13 May 2013}
\availableonline{15 May 2013}
\communicated{S. Sarkar}






\begin{abstract}
Recognizing objects and scenes are two challenging but essential tasks in image understanding. In particular, the use of RGB-D sensors in handling these tasks has emerged as an important area of focus for better visual understanding. Meanwhile, deep neural networks, specifically convolutional neural networks (CNNs), have become widespread and have been applied to many visual tasks by replacing hand-crafted features with effective deep features. However, it is an open problem how to exploit deep features from a multi-layer CNN model effectively. In this paper, we propose a novel two-stage framework that extracts discriminative feature representations from multi-modal RGB-D images for object and scene recognition tasks. In the first stage, a pretrained CNN model has been employed as a backbone to extract visual features at multiple levels. The second stage maps these features into high level representations with a fully randomized structure of recursive neural networks (RNNs) efficiently. To cope with the high dimensionality of CNN activations, a random weighted pooling scheme has been proposed by extending the idea of randomness in RNNs. Multi-modal fusion has been performed through a soft voting approach by computing weights based on individual recognition confidences (i.e. SVM scores) of RGB and depth streams separately. This produces consistent class label estimation in final RGB-D classification performance. Extensive experiments verify that fully randomized structure in RNN stage encodes CNN activations to discriminative solid features successfully. Comparative experimental results on the popular Washington RGB-D Object and SUN RGB-D Scene datasets show that the proposed approach achieves superior or on-par performance compared to state-of-the-art methods both in object and scene recognition tasks. Code is available at \url{https://github.com/acaglayan/CNN_randRNN}.
\end{abstract}




\end{frontmatter}


\section{Introduction}
\label{sec:intro}
\begin{figure*}[!t]
	\centering
	\includegraphics[width=0.85\textwidth]{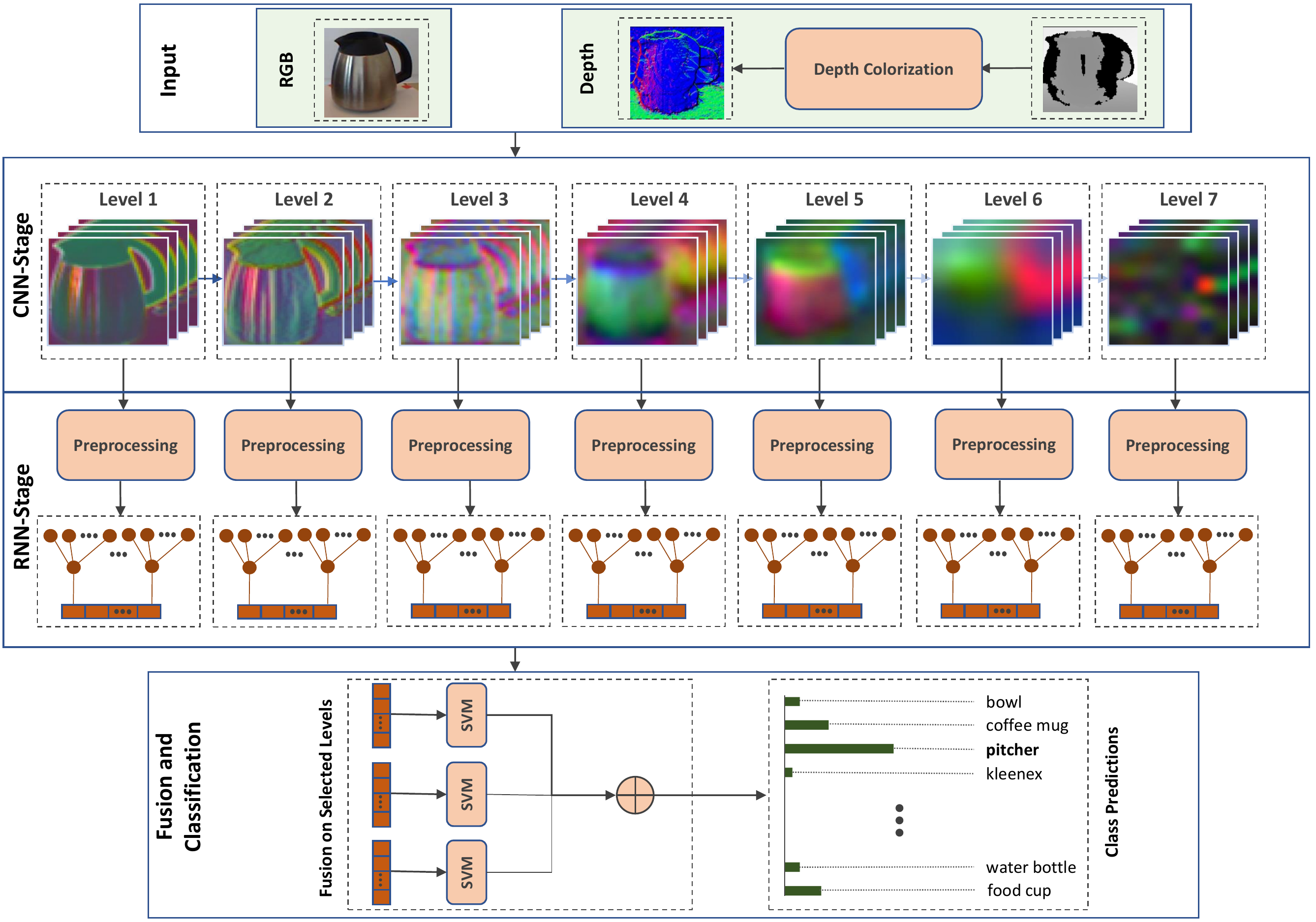}
	\caption{General overview of the proposed framework. The framework accepts RGB and depth images as inputs. In the CNN-Stage, activations at different levels of a pretrained model are extracted. In the RNN-Stage, first, CNN activations are adjusted through preprocessing operations so that they can be utilized by RNNs to output same number of feature vectors at each level. Then, multiple random RNNs are applied to map these inputs into high level representations. Finally, multiple level fusion and classification steps are deployed for recognition tasks.}
	\label{fig:FrameworkStructure}
\end{figure*}
Convolutional neural networks (CNNs) have attracted researchers to handle many visual recognition tasks since their breakthrough emergence. However, building an effective model can be quite challenging due to the lack of labeled training data, limited time and computational resources, and the need for well defined hyperparameter settings for a good generalization capability. Especially in many real-world tasks, it is not preferable to train a model from scratch. Luckily, CNNs offer highly efficient solutions with their transferable off-the-shelf features. Consequently, many approaches take advantage of these features to propose new solutions for object recognition (e.g. \cite{Razavian_CVPRW_2014, Schwarz_ICRA_2015}), scene recognition (e.g. \cite{Liao_ICRA_2016, Song_TIP_2019}), object detection (e.g. \cite{Girshick_CVPR_2014, Sermanet_ICLR_2014}), and semantic segmentation (e.g. \cite{Girshick_CVPR_2014, Farabet_TPAMI_2013}) due to their high representation ability and capability of generalization among different tasks when trained with large scale datasets. The most common and straightforward strategy among these methods is to utilize the features obtained from final layers which provide semantically rich information with smaller dimensions comparing to the earlier layers \citep{Razavian_CVPRW_2014, Schwarz_ICRA_2015, Girshick_CVPR_2014, Sermanet_ICLR_2014, Farabet_TPAMI_2013}. However, one of the concerns about this semantics is the fact that as features evolve towards the final layers, they are increasingly dependent on the chosen dataset and task \citep{Yosinski_NIPS_2014}, which might diminish the generalization capabilities of these features when transferred. Moreover, this strategy ignores the locally activated distinctive information of the earlier layers which is less sensitive to semantics \citep{Hariharan_CVPR_2015, Zaki_ICRA_2016}. One of the main challenges in earlier layers of deep CNNs is the high dimensionality of extracted features. In addition, when these features are used as is, it makes the feature space untraceable. Eventually, while features are transformed from low-level general to high-level specific representations throughout the network, the relational information is distributed across the network at different levels \citep{Yosinski_NIPS_2014, Hariharan_CVPR_2015}. However, it remains unclear how to exploit the information effectively.

\textbf{Motivation and Proposed Work:} In this paper, we aim to present an effective deep feature extraction framework to derive powerful image representations through transfer learning. The proposed pipeline relies on two key insights. The first one is to employ a pretrained CNN as the backbone model and exploit activations at different layers of the network to cover the predominant information of the underlying localities. The second one is to implement multiple random recursive neural networks (RNNs) on top of CNNs to encode the CNN activations into a robust representation with reduced dimensionality and sufficient descriptiveness. Our motivation is twofold. We want our framework to generate highly discriminative deep feature representations without the need for training during the feature extraction stage, yet provide a training capability for extra performance boost. We question whether a fully random neural network stage lacks representation power.

In developing our framework, we particularly deal with the RGB-D object and scene recognition problems, which are challenging yet crucial tasks especially with the today’s wider application of robotics technologies. Moreover, the multi-modality of the RGB-D sensors arises additional difficulties in representation of input data such as handling different modalities and devising solutions that captures complementary information from both RGB and depth data effectively. Besides these challenges, alleviating limitations on time and memory consumption is another challenge to deal with. To address these challenges, we propose a novel framework that gathers feature representations at different levels in a compact and representative feature vector for both of RGB and depth data. After obtaining CNN activations, we first apply a preprocessing operation to the activation maps of each level through reshaping or randomized pooling. This not only provides a generic structure for each level by fixing an RNN tree but also it allows us to improve recognition accuracy through multi-level fusion. We then give the outputs of these operations to multiple random RNNs \citep{Socher_NIPS_2012} to acquire higher level compact feature representations. Incorporating multiple fixed RNNs together with the pretrained CNN models allows feature transition at different levels to preserve both semantic and spatial structure of objects. In order to transfer learning from a pretrained CNN model for depth modality, we embed depth data into the RGB domain with a highly efficient depth colorization technique based on surface normals. As for the multi-modal fusion of RGB and depth modalities, we explore different fusion techniques. Moreover, we present an approach that provides a decisive fusion of RGB and depth modalities based on the modality importance through a weighting scheme (see Sec. \ref{sec:fusionClassification}).  

The proposed framework is evaluated with exhaustive experiments on two popular public datasets (i) Washington RGB-D Object dataset \citep{Lai_ICRA_2011} for RGB-D object recognition task and (ii) Sun RGB-D Scene dataset \citep{Song_CVPR_2015} for RGB-D scene recognition task. The experimental results demonstrate the effectiveness of our approach in terms of accuracy by achieving superior performance over the current state-of-the-art methods. A preliminary version of this work appeared in \citep{Caglayan_ECCVW_2018} for RGB-D object recognition. In our preliminary research, we have already explored various properties of RNNs, such as non-linearity functionality, comparative accuracy performance and feature size over the use of CNN-only features. In this work, we present an extended and enhanced version of our work in \citep{Caglayan_ECCVW_2018} with a novel framework and make the following improvements. First, we improve the idea by designing and implementing the pipeline from scratch. 
Second, we have made the proposed work applicable to a  variety of backbone models from shallow to deep. To this end, we introduce a random pooling strategy as a preprocessing step to deal with the high dimensionality of the activation maps of deep models such as ResNet and DenseNet, so that early layers of these models could be utilized in our random recursive neural networks. Third, we introduce a soft voting approach for multi-modal RGB-D fusion based on individual classification confidences of each modality. This provides better accuracy performance in recognition tasks. Fourth, although prior research has presented randomness in neural networks with various approaches such as feature extraction systems in \citep{Jarrett_ICCV_2009, Socher_NIPS_2012, Cheng_CVIU_2015, Bui_Access_2016} and stochastic pooling \citep{Zeiler_ICLR_2013}, in this work, we improve on these and elaborate randomness both in technical perspective and empirical investigation. Finally, we extend the proposed approach to RGB-D scene recognition task and achieve the state-of-the-art results in challenging benchmarks.

\textbf{Contributions:} To sum up, the main contributions of this paper can be listed as follows:
\begin{itemize}
	\item We present a novel framework (see Fig.~\ref{fig:FrameworkStructure}) for deep features with two-stage organization where information at different levels is encoded by incorporation of multiple random RNNs with a pretrained CNN model for RGB-D object and scene recognition (see Sec. \ref{sec:method}). The framework is applicable to a variety of pretrained CNN models including AlexNet \citep{Krizhevsky_NIPS_2012}, VGGNet \citep{Simonyan_ICLR_2015}, ResNet \citep{He_CVPR_2016}, and DenseNet \citep{Huang_CVPR_2017}. The overall structure has been designed in a modular and extendable way through a unified CNN and RNN process. Thus, it offers easy and flexible use. These also can easily be extended with new capabilities and combined with different setups and other models for implementing new ideas. In fact, our preliminary approach has been already successfully applied to another challenging robotics task in a SLAM system \citep{Guclu_CVPRW_2019}.
	
	\item We extend the idea of randomness in RNNs as a novel pooling strategy to cope with the high dimensionality of CNN activations from different levels (see Sec. \ref{sec:randomPooling}). This strategy has been applied as a preprocessing stage before RNNs and it allows us to evaluate and utilize multiple level information in deep models such as ResNet \citep{He_CVPR_2016} and DenseNet \citep{Huang_CVPR_2017} models. In addition, we give the experimental results of different pooling strategies in terms of accuracy and show the effectiveness of our pooling strategy over other pooling methods (see Sec. \ref{sec.exp.ma.poolingPerformances}).

	\item We study several aspects of transfer learning through an empirical investigation including level-wise analysis of different baselines and the effects of finetuning over fixed pretrained CNN models (see Sec. \ref{sec:exp.modelAblation} and the \textit{supplementary material} for further experimental analysis). In regard to multi-model fusion, unlike our previous work using concatenation of features, we propose a soft voting approach based on individual SVM confidences of RGB and depth streams (see Sec. \ref{sec:fusionClassification}) and show the strength of our approach experimentally (see Sec. \ref{sec.exp.ma.weightedFusion}). We also provide experimental results demonstrating that our approach improves the state-of-the-art results on two challenging real-world public datasets: Washington RGB-D Object dataset for RGB-D object recognition (see Sec. \ref{sec:exp.objectRecognition}) and SUN RGB-D scene dataset for RGB-D scene recognition (see Sec. \ref{sec:exp.sceneRecognition}).
\end{itemize}

\section{Related Work}
\label{sec:relatedwork}
The proposed work can be related with different areas, such as multi-modal CNN based approaches, transfer learning based approaches, and random recursive neural networks. In this section, we narrow our focus to RGB-D based recognition and give a brief review of the relevant approaches with stating the current work in the literature.

\subsection{Multi-Modal CNN based Approaches}
Following their success in computer vision, CNN-based solutions have replaced conventional methods such as the works in \cite{DepthKernel2011}, \cite{HKD2011}, and \cite{HONV2012} in the field of RGB-D object recognition, as in many other areas. For instance, the authors of \citep{Wang_2015_ICCV, Wang_2015_IEEE_ToM} present CNN-based multi-modal learning systems motivated by the intuition of common patterns shared between RGB and depth modalities. They enforce their systems to correlate features of the two modalities in a multi-modal fusion layer with a pretrained model \citep{Wang_2015_ICCV} and their custom network \citep{Wang_2015_IEEE_ToM} respectively. \cite{Li_AAAI_2018df} extend the idea of considering multi-modal intrinsic relationship with intra-class and inter-class similarities for indoor scene classification by providing a two-stage training approach. In \cite{Rahman_ICME_2017}, a three-streams multi-modal CNN architecture has been proposed in which depth images are represented with two different encoding methods in two-streams and the remaining stream is used for RGB images. Despite the extra burden, this naturally has increased the depth accuracy in particular. Similar multi-representational approach has been proposed by \cite{Zia_ICCVW_2017} where a hybrid 2D/3D CNN model initialized with pretrained 2D CNNs is employed together with 3D CNNs for depth images. \cite{Cheng_3DV_2015} propose convolutional fisher kernel (CFK) method which integrates a single CNN layer with fisher kernel encoding and utilizes Gaussian mixture models for feature distribution. The drawback of their approach is the very high dimensional of the feature space.

\subsection{Transfer Learning based Approaches}
Deep learning algorithms require a significant amount of annotated training data and obtaining such data can be difficult and expensive. Therefore, it is important to leverage transfer learning for enhancing high-performance learner on a target domain and the task at hand. Especially, applying a trained deep network and then fine-tuning the parameters can speed up the learning process or improve the classification performance \citep{Wang_TIP_2017}. Furthermore, many works show that a pretrained CNN on a large-scale dataset can generate good generic representations that can effectively be used for other visual recognition tasks as well \citep{Razavian_CVPRW_2014, Yosinski_NIPS_2014, Oquab_CVPR_2014, Azizpour_CVPRW_2015, Azizpour_TPAMI_2015}. This is particularly important in vision tasks on RGB-D datasets, which is hard to collect with labeled data and generally amount of data is much less than that of the labeled images in RGB datasets.

There are many successful approaches that use transfer learning in the field of RGB-D object recognition. \cite{Schwarz_ICRA_2015} use the activations of two fully connected layers, i.e. \textit{fc7} and \textit{fc8}, extracted from the pretrained AlexNet \citep{Krizhevsky_NIPS_2012} for RGB-D object recognition and pose estimation. \cite{Gupta_ECCV_2014} study the problem of object detection and segmentation on RGB-D data and present a depth encoding approach referred as HHA to utilize a pretrained CNN model on RGB datasets. Asif \textit{et al.} introduce a cascaded architecture of random forests together with the use of the \textit{fc7} features of the pretrained models of \citep{Chatfield_BMVC_2014} and \citep{Simonyan_ICLR_2015} to encode the appearance and structural information of objects in their works of \cite{Asif_ICRA_2015} and \cite{Asif_ToR_2017}, respectively. \cite{Carlucci_RAS_2018} propose a colorization network architecture and use a pretrained model as feature extractor after fine-tuning it. They also make use of the final fully-connected layer in their approach. So, these above-mentioned studies mainly focus on the outputs of the fully-connected layers. 

On the other hand, many studies \citep{Liu_CVPR_2015, Zaki_ICRA_2016, Zaki_RAS_2017, Song_IJCAI_2017, Caglayan_ECCVW_2018} have concluded that using fully connected layers from pretrained or finetuned networks might not be the optimum approach to capture discriminating properties in visual recognition tasks. Moreover, combining the activations obtained in different levels of the same modal enhances recognition performance further, especially for multi-modal representations, where earlier layers capture modality-specific patterns \citep{Yang_ICCV_2015, Song_IJCAI_2017, Caglayan_ECCVW_2018}. Hence, utilizing information at different levels in the works of \citep{Yang_ICCV_2015, Zaki_ICRA_2016, Zaki_RAS_2017, Song_IJCAI_2017, Caglayan_ECCVW_2018, Zaki_AuotRobots_2019} yields better performances. More recent approach of \cite{Loghmani_RAL_2019} utilizes the pretrained model of residual networks \citep{He_CVPR_2016} to extract features from multiple layers and combines them through a recurrent neural network. Their experimental results also verify that multi-level feature fusion provides better performance than single-level features. While their approach is based on a gated recurrent unit (GRU) \citep{Cho_EMNLP_2014} with a number of memory neurons, our approach employs multiple random neural networks with no necessarily need for training. A different related approach is proposed by \cite{Asif_TPAMI_2018}. They handle the classification task by dividing it into image-level and pixel-level branches and fusing through a Fisher encoding branch. \cite{Eitel_IROS_2015} and \cite{Tang_TCDS_2019} employ two-stream CNNs, one for each modality of RGB and depth channels and each stream uses the pretrained model of \citep{Krizhevsky_NIPS_2012} on the ImageNet. In both works \citep{Eitel_IROS_2015, Tang_TCDS_2019}, the two-streams are finally connected by a fully-connected fusion layer and a canonical correlation analysis (CCA) module, respectively. While feature fusion approaches (e.g. concatenation) may provide good accuracy for the visual recognition task, feature fusion may not be the only solution for multi-level decision process since increased feature space may not be good for recognition with small number of data. We experiment and show that voting on the SVM confidence scores for selected levels can also provide reliable and improved performance. Moreover, this also enables us to use confidence score based importance to RGB and depth domains in multi-modal fusion.

\subsection{Random Recursive Neural Networks}
Randomization in neural networks has been researched for a long time in various studies \citep{Schmidt_ICPR_1992, Pao_Computer_1992, Pao_Neurocomp_1994, Igelnik_Pao_TNN_1995, Huang_TNN_2006, Rahimi_NIPS_2008, Socher_NIPS_2012} due to its benefits, such as simplicity and computationally cheapness over optimization \citep{Rahimi_NIPS_2009}. Since a complete overview of these variations is beyond the scope of this paper, we give an overview specifically with the focus of random recursive neural networks \citep{Socher_NIPS_2012}. Recursive neural networks (RNNs) \citep{Pollack_AI_1990, Hinton_AI_1990, Socher_ICML_2011} are graphs that process a given input into recursive tree structures to make a high-level reasoning possible in a part-whole hierarchy by repeating the same process over the trees. RNNs have been employed for various research purposes in computer vision including image super-resolution \citep{Kim_CVPR_2016}, semantic segmentation \citep{Socher_ICML_2011, Sharma_NIPS_2014}, and RGB-D object recognition \citep{Socher_NIPS_2012, Bai_Neurocomp_2015, Cheng_CVIU_2015}. \cite{Socher_NIPS_2012} have introduced a two-stage RGB-D object recognition architecture where the first stage is a single CNN layer using a set of k-means centroids as the convolution filters and the second stage is multiple random recursive neural networks to process outputs of the first stage. \cite{Bai_Neurocomp_2015} propose a subset based approach of the pioneer work in \citep{Socher_NIPS_2012} where they use a sparse auto-encoder instead of the k-means clustering for convolution filters. \cite{Cheng_CVIU_2015} employ the same architecture of \cite{Socher_NIPS_2012} for a semi-supervised learning system with a modification by adding a spatial pyramid pooling to prevent a potential performance degradation during resizing input images. \cite{Bui_Access_2016} have replaced the single CNN layer in \citep{Socher_NIPS_2012} with a pretrained CNN model for RGB object recognition and achieved impressive results. Following their success, in our preliminary work \citep{Caglayan_ECCVW_2018}, we propose an approach that aims to improve on this idea by gathering feature representations at different levels in a compact and representative feature vector for both of RGB and depth data. To this end, we reshape CNN activations in each layer that provides a generic structure for each layer by fixing the tree structure without hurting performance and it allows us to improve recognition accuracy by combining feature vectors at different levels. In this work,  we propose a pooling strategy to handle large dimensional CNN activations by extending the idea of randomness in RNNs. This can be related with the stochastic pooling in \cite{Zeiler_ICLR_2013}, which picks the normalized activations of a region according to a multinomial distribution by computing the probabilities within the region. Instead of using probabilities, our pooling approach here is a form of averaging based on uniform distributed random weights.

\section{Proposed Approach}
\label{sec:method}
The proposed pipeline has two main stages. In the first stage, a pretrained CNN model has been employed as the underlying feature extractor. In this work, we have examined several models in this stage. The second stage transforms convolutional features through a randomized recursive neural network based structure that aims to acquire more compact representations. To cope with the high dimensionality of CNN activations, a pooling strategy based on random weights has been proposed. The final representative outcomes have been passed through a linear SVM classifier for categorization of objects and scenes. The overall pipeline can be related as a deeper analogy to \cite{Jarrett_ICCV_2009} where a proper architecture with random weights for object recognition task has been explored. 

In order to use pretrained CNN models, it is important to process input images appropriately. To this end, we perform a set of data preparation processes on both RGB and depth data and represent depth data into an effective 3-channel structure similar to RGB data using surface normal estimation (see the \textit{supplementary material} for details).

\subsection{CNN-Stage}
The backbone of our approach is a pretrained CNN model. Since size of available RGB-D datasets are much smaller than that of RGB's, it is important to make use of an efficient knowledge transfer from pretrained models on large RGB datasets. In addition, it saves time by eliminating the need for training from scratch. In the previous work \citep{Caglayan_ECCVW_2018}, the available pretrained CNN model of \citep{Chatfield_BMVC_2014}, named VGG\_f, in MatConvNet toolbox \citep{Vedaldi_Matconvnet_ICM_2015} has been used. In this work, we employ several available pretrained models of the ImageNet including AlexNet \citep{Krizhevsky_NIPS_2012}, VGGNet \citep{Simonyan_ICLR_2015} (specifically VGGNet-16 model with batch normalization), ResNet \citep{He_CVPR_2016} (specifically ResNet-50 and ResNet-101 models), and DenseNet \citep{Huang_CVPR_2017}. We extract features from seven different levels of CNN models. For AlexNet, outputs of the five successive convolutional layers and the following two fully-connected (FC) layers have been considered, while for VGGNet, the first two FC layers  are taken into account together with the outputs of each convolution block that includes several convolutions and a final max pooling operations. Unlike AlexNet and VGGNet, ResNet and DenseNet models consist of blocks such as residual, dense or transition blocks where there are multiple layers. While ResNet extends the sequential behaviour of AlexNet and VGGNet with the introduction of the skip-connections, DenseNet takes one step further by concatenating the incoming activations rather than summing up them. The ResNet models consist of five stages and a following average pooling and an FC layer. Therefore, each output of the five successive stages and the output of the final average pool have been considered for the six of the seven extraction points. As for the remaining extraction level for these models (ResNet-50 and ResNet-101), the middle point of the third block (which is the largest block) has been taken. Similarly, for DenseNet model, the output of all the four dense blocks (for the last dense block, the output of normalization that follows the dense block has been taken) and the transition blocks between them have been considered as the extraction points. Since common and straightforward model of AlexNet has a minimum depth with a seven layer stack-ups, the above-mentioned CNN extraction points for each model are selected to evaluate and compare level-wise model performances. In addition, these levels are also related to the CNN model in the previous work \citep{Caglayan_ECCVW_2018} that we improve on by considering their intrinsic reasoning behind the use of blocks and the approximate distance differences.

\subsection{RNN-Stage}
\begin{figure}[!t]
	\centering
	\includegraphics[width=0.65\columnwidth, keepaspectratio]{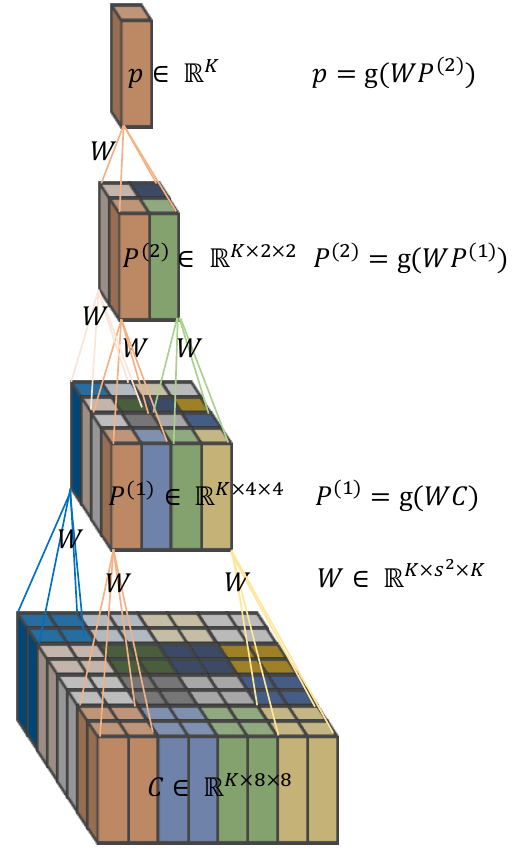}
	\caption{Graphical representation of a single recursive neural network (RNN). The same random weights have been applied to compute each node and level.}
	\label{fig:RNN}
\end{figure}
Random recursive neural networks offer a feasible solution by randomly fixing the network connections and eliminate the need for selection in the parameter space. Motivated by this, we employ multiple random RNNs, whose inputs are the activation maps of a pretrained CNN model. RNNs map a given 3D matrix input into a vector of higher level representations of it by applying the same operations recursively in a tree structure. In each layer, adjacent blocks are merged into a parent vector with tied weights where the objective is to map inputs $\mathit{C \in}$ $\mathbb{R}$$\mathit{^{K \times s \times s}}$ into a lower dimensional space $\mathit{p \in}$ $\mathbb{R}$$\mathit{^{K}}$ through multiple levels in the end. Then, the output of a parent vector is passed through a nonlinear function. A typical choice for this purpose is the $\mathit{tanh}$ function. In our previous work \citep{Caglayan_ECCVW_2018}, we give the comparative results of different activation functions in terms of accuracy success and show hyperbolic functions work well. Therefore, in this work, we employ $\mathit{tanh}$ activation function as in \citep{Socher_NIPS_2012, Caglayan_ECCVW_2018}. Fig. \ref{fig:RNN} shows a graphical representation of a pooled CNN output with the size $K\times8\times8$ and an RNN structure with 3 levels and blocks of $2\times2=4$ child nodes (Note that this figure is inspired by the RNN graphical representation of \cite{Socher_NIPS_2012}).

In our case, inputs of RNNs are activation maps obtained from different levels of the underlying CNN model. Let $\mathit{x}$ be an input image that pass through $\mathit{f(x)_l}$ a given CNN model, where $\mathit{l=1,..,7}$ are the extraction levels and $\mathit{f(x)_l = C_l}$, where the output convolution maps are either a 3D matrix $\mathit{C_l \in \mathbb{R}^{K\times s\times s} }$ for $\mathit{l}$ convolutional layers or a 1D vector of $\mathit{C_l \in \mathbb{R}^{M} }$ for $\mathit{l}$ FC layers/global average pooling. Since RNN requires a 3D input of $\mathit{C \in \mathbb{R}^{K \times s \times s}}$, we first process the convolution maps at each level to ensure the required form. Moreover, by applying this step, we ensure that RNNs are able to handle inputs fast and effectively by reducing the receptive field area and/or the number of activation maps of high-dimensional feature levels (e.g. the outputs of early levels for models such as VGGNet, ResNet, DenseNet, etc.). In addition, we apply preprocessing to obtain similar output structures with the previous work \cite{Caglayan_ECCVW_2018}. However, it was enough to apply only reshaping in the previous work due to less dimensional size of layers in VGG\_f model. In this work, we introduce random weighted pooling that copes with high dimensionality of layers in the underlying deeper models such as ResNet \citep{He_CVPR_2016} and DenseNet \citep{Huang_CVPR_2017}. Our pooling mechanism can downsample CNN activations in both number and spatial dimension of maps. After applying the preprocessing step to obtain suitable forms for RNNs, we compute parent vector as
\begin{equation} \label{eq:rnnParent}
  \begin{aligned}
      p = g\left(WC_l\right)\\
  \end{aligned}
\end{equation}
where $\mathit{C_l = \begin{bmatrix} c_1 \\ \vdots \\ c_{s^2} \end{bmatrix}}$ for each CNN extraction level $\mathit{l=1,...,7}$, $g$ is a nonlinearity function which is $tanh$ in this study, $s$ is block size of an RNN. Instead of a multi-level structured RNN, an RNN in this study is of one-level with a single parent vector. In fact, our experiments have shown that the single-level structure provides better or comparable results over the multi-level structure in terms of accuracy (see the \textit{supplementary material}). Moreover,  the single-level is more efficient with less computational burden. Thus, $s$ block size is actually the receptive field size in an RNN. In Eq. \ref{eq:rnnParent}, the parameter weight matrix is $\mathit{W \in \mathbb{R}^{K\times s^2K}}$ and it is randomly generated from a predefined distribution that satisfies the following probability density function
\begin{equation} \label{eq:weightPdf} 
  \begin{aligned}
      W \sim h \Rightarrow \int_{a}^{b}h(w)dw = P(a \leq W \leq b)\\
  \end{aligned}
\end{equation}
where $h$ is a predefined distribution and $a$ and $b$ are boundaries of the distribution. In our case, the weights are set to be uniform random values in $\mathit{[-0.1, +0.1]}$, which have been assigned by following our previous work \cite{Caglayan_ECCVW_2018} and specifically with the assumption of preventing possible explosion of tensor values due to our aggregating pooling strategy. On the other hand, \cite{Saxe_ICML_2011} find that the distribution of random weights such as uniform, Laplacian, or Gaussian does not affect classification performances as long as the distribution is 0-centered. We refer readers to \cite{Rahimi_NIPS_2008} and \cite{Rudi_NIPS_2017} for more insights and further details on the properties of random features. Keeping in mind that in order to obtain sufficient descriptive power from the randomness, we need to generate enough samples from the range. In \cite{Socher_NIPS_2012}, it has been demonstrated experimentally that increasing the number of random RNNs up to $64$ improves performance and gives the best result with $128$ RNNs. In \cite{Caglayan_ECCVW_2018}, it has also been verified that $K = 128$ number of RNN weights can be generated for feature encoding with high performance in classification on both of RGB and depth data. Therefore, as a standard usage in this work, we do feature encoding on CNN features using $128$ random RNNs with $64$ channel representations, leading us to $8192$ dimensional feature vector at each level in a model.

The reason why random weights work well for object recognition tasks seems to lie in the fact that particular convolutional pooling architectures can naturally produce frequency selective and translational invariant features \citep{Saxe_ICML_2011}. As stated before, in analogy to the convolutional-pooling architecture in \cite{Jarrett_ICCV_2009}, our approach intuitively incorporates both selectivity due to the CNN stage and translational invariance due to the RNN stage. Moreover, we have to point out that there is biological plausibility lies in the use of randomness as well. \cite{Rigotti_Frontiers_2010} have shown that random connections between inter-layer neurons are needed to implement mixed selectivity for optimal performance during complex cognitive tasks. Before concluding this section, we give details of our random pooling approach, where we extend the idea of random RNN as a downsampling mechanism.

\subsubsection{Random Weighted Pooling} \label{sec:randomPooling}
\begin{figure}
	\centering
	\includegraphics[width=\columnwidth, keepaspectratio]{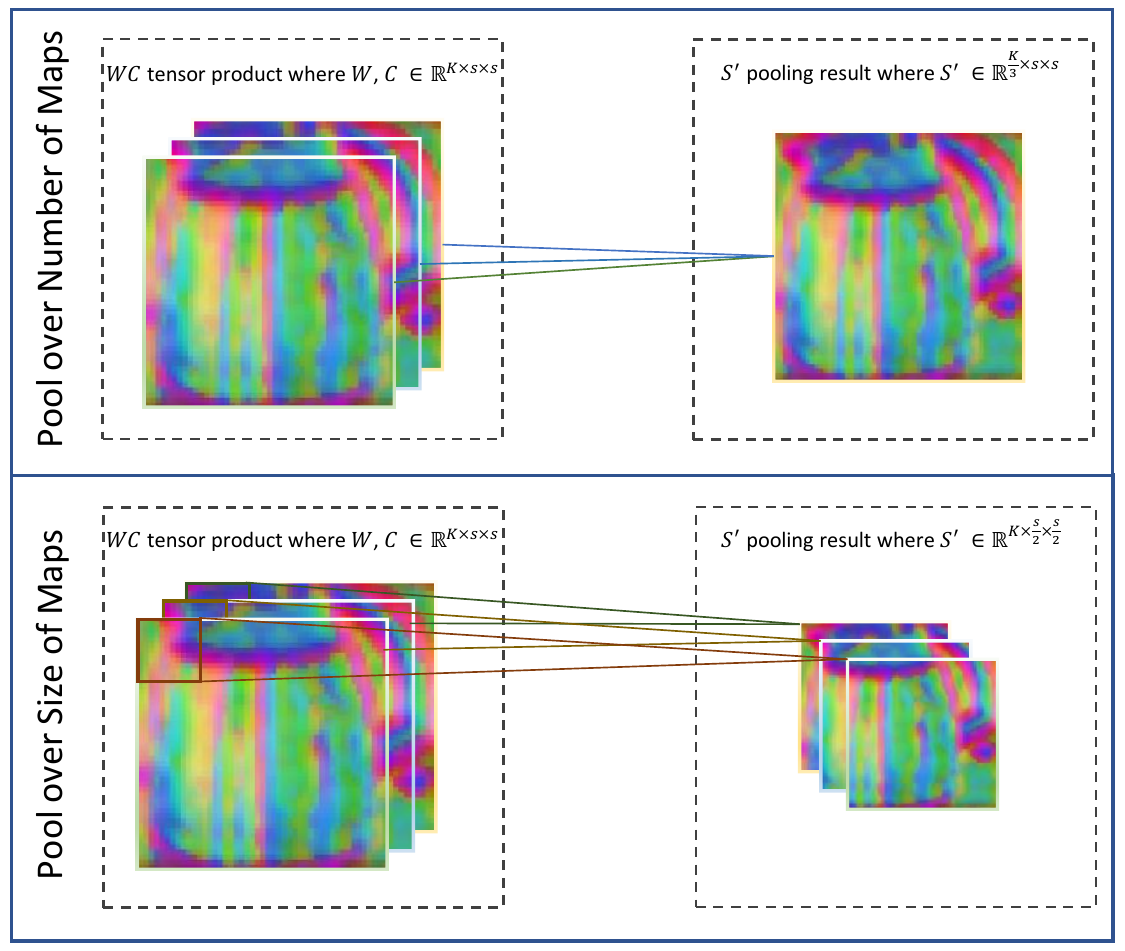}
	\caption{Illustration of random weighted pooling over number of maps (top) and window size of maps (below).}
	\label{fig:Pooling}
\end{figure}

In our previous work \citep{Caglayan_ECCVW_2018}, we give CNN outputs to RNNs after a reshaping process. However, due to the high dimensional output size of the models used in this study, it is necessary to process CNN activations further. In this work, we propose a random pooling strategy to reduce the dimension in either size of the activation maps ($s$ block size or receptive field area of an RNN) or number of maps ($K$) at CNN levels where reshaping is insufficient. In our random weighted pooling approach, we aggregate the CNN activation maps by sampling from a uniform distribution as in Eq. \ref{eq:weightPdf} from each pooling area. More precisely, for $l$ extraction level, the pooling reduces $\mathit{C_l}$ activations by mapping into $\mathit{A_{l}^{'}}$ area as $\mathit{P: C_l \mapsto A_{l}^{'}}$ where $\mathit{C_l \in \mathbb{R}^{K\times s \times s}}$ and $\mathit{A_{l}^{'} \in \mathbb{R}^{K^{'}\times s^{'} \times s^{'}}}$ in Eq. \ref{eq:randomPool}.
\begin{equation} \label{eq:randomPool}
  \begin{aligned}
      A_l^{'} = \sum_{i \in A_l}W_{l}^{(i)}{C_{l}^{(i)}}\\
  \end{aligned}
\end{equation}
where $A_l$ is pooling area, $C_l$ convolutional activations, $i$ is the index of each element within the pooling, and $W_l$ is random weights. $K^{'}< K$ and $s^{'} = s$ when pooling is over number of maps whereas $K^{'} = K$ and $s^{'} < s$ when pooling is over size of maps. Fig. \ref{fig:Pooling} illustrates proposed random weighted pooling for both of downsampling in number of maps and size of maps. In this work, by extending the randomness in RNNs along the pipeline with the proposed pooling strategy, we aim to show that randomness can actually work quite effectively. In fact, as we can see in the comparative results (see Sec. \ref{sec.exp.ma.poolingPerformances}), this randomness in our approach works generally better comparing to the other common pooling methods such as max pooling and average pooling, especially at the semantic levels.
\subsection{Fusion and Classification} \label{sec:fusionClassification}
After obtaining encoded features from the RNN-Stage, we investigate multi-level fusions to capture more distinctive information at different levels for further recognition performance. To minimize the cross entropy error between output predictions and the target values, we could give multi-level outputs to fully connected layers and back-propagate through them. However, following the success in our previous study \citep{Caglayan_ECCVW_2018}, we perform classification by employing linear SVM with the scikit-learn\footnote{\url{https://github.com/scikit-learn/scikit-learn}} \citep{Pedregosa_JMLR_2011_scikit} implementation. To this end, in our previous work \citep{Caglayan_ECCVW_2018}, we have performed the straightforward feature concatenation on various combinations of the best mid-level representations. In this work, in addition to the feature concatenation, we also apply soft voting by averaging SVM confidence scores on these best trio of levels. Finally, RGB and depth features are fused to evaluate combined RGB-D accuracy performance. 

The motivation behind the need for a complementary multi-modal fusion is twofold. The fact that shiny, transparent, or thin surfaces may cause corruption in depth information since depth sensors do not properly handle reflections from such surfaces, resulting better performance in favor of RGB in such cases. On the other hand, depth sensors work well in a certain range and are insensitive to changes in lighting conditions. Therefore, to take full advantage of both modalities in a complementary way, a compact multi-modal combination based on the success of input type is important in devising the best performing fusion. To this end, we present a decision mechanism using weighted soft voting based on the confidence scores obtained from RGB and depth streams. Modality weighting in this way is used to compensate imbalance and complement decision in different data modalities. Once the modality-specific branches proceed, we combine the predictions through the weighted SVM as follows. Let $S_{\scriptstyle{i}}$ represents SVM confidence scores of each category class $n = 0...N-1$, where $N$ is number of classes, and $i \in \{rgb, depth\}$ indicates RGB and depth modalities. Then, weights $w_{\scriptstyle{i}}$ are computed as in Eq. \ref{eq:fusionWeights}. 
\begin{equation} \label{eq:fusionWeights}
  \begin{aligned}
     w_{\scriptstyle{i}} = \sqrt{\dfrac{e^{m_{\scriptstyle{i}}}}{\sum\limits_{i} {e^{m_{\scriptstyle{i}}}}}}
  \end{aligned}
\end{equation}
where $m_{\scriptstyle{i}}$ is normalized squared magnitudes for each modality and defined as:
\begin{equation} \label{eq:normScoreMagnitudes}
  \begin{aligned}
     m_{\scriptstyle{i}}=\dfrac{{\lVert S_{\scriptstyle{i}}\rVert}^2}{\max({\lVert S_{\scriptstyle{rgb}}\rVert}^2, {\lVert S_{\scriptstyle{depth}}\rVert}^2)}
  \end{aligned}
\end{equation}

Finally, multi-modal RGB-D predictions are estimated as follows, in Eq. \ref{eq:weightedFusion}:
\begin{equation} \label{eq:weightedFusion}
  \begin{aligned}
      \hat{y}_{\scriptscriptstyle{RGBD}} = {\underset{n}{\mathrm{arg\,max}}} \sum_i w_{\scriptstyle{i}}S_{\scriptstyle{i}}
  \end{aligned}
\end{equation}
where $n$ is a category class. Concretely, if RGB and depth results are balanced in confidence scores, then the final soft voting decision is based on equal contribution from each stream similar to averaging.

\section{Experimental Evaluation}
\label{sec:experiments}
The proposed framework has been evaluated on two challenging benchmarks for two tasks: (i) RGB-D object recognition (Sec. \ref{sec:exp.objectRecognition}) using Washington RGB-D object dataset \citep{Lai_ICRA_2011} and (ii) RGB-D scene recognition (Sec. \ref{sec:exp.sceneRecognition}) using SUN RGB-D scene dataset \citep{Song_CVPR_2015}. After introducing the datasets and setups, we first compare our results with state-of-the-art results for both benchmarks. Results of other methods are taken from the original papers. Then, we carry out extensive experiments (Sec. \ref{sec:exp.modelAblation}) on the challenging Washington RGB-D object dataset, which is a larger-scale RGB-D dataset comparing to other RGB-D benchmarks, to evaluate effects of various model parameters and setup properties in our framework.
\begin{figure*}[!ht]
	\begin{center}
		\includegraphics[width=\textwidth, keepaspectratio]{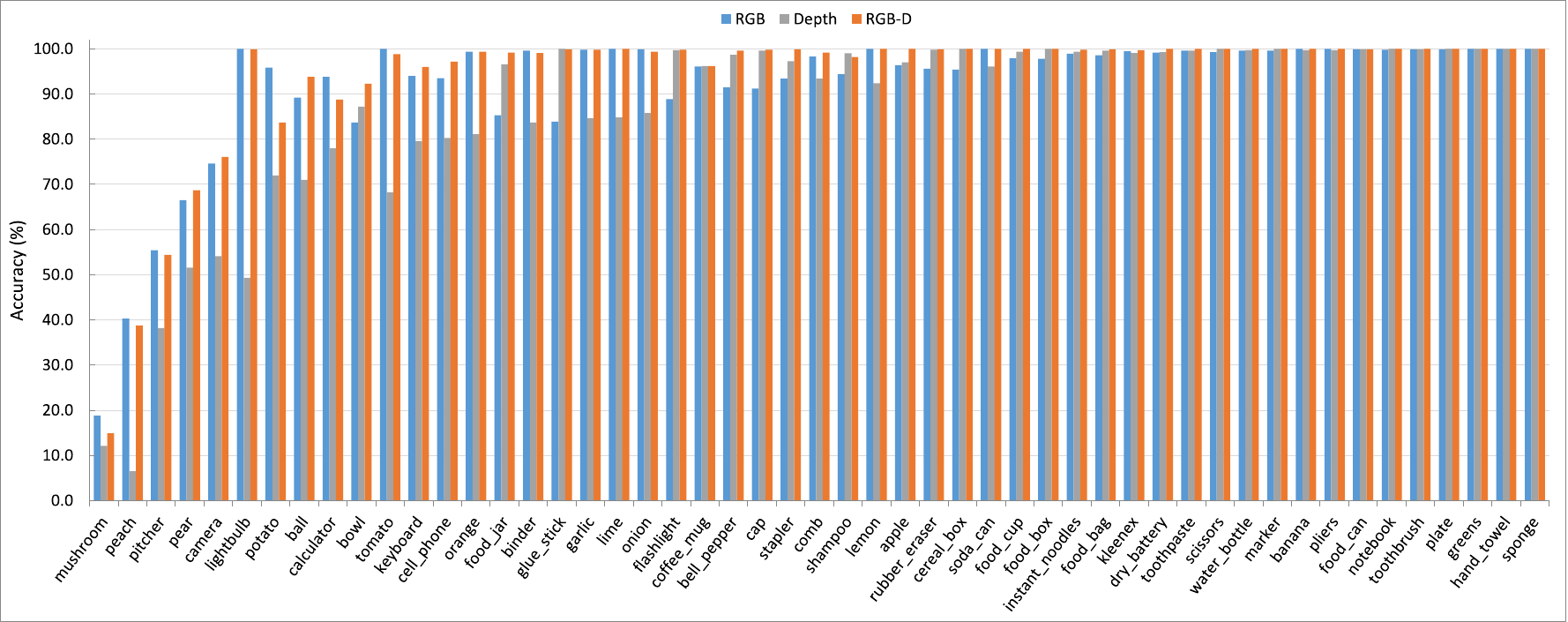}
	\end{center}
	\caption{Per-category average accuracy performances of ResNet101-RNN on Washington RGB-D Object dataset.}
	\label{fig:wrgbdIndividualResults}
\end{figure*}
\subsection{Dataset and Setup} \label{sec:exp.datasets}
\subsubsection{Washington RGB-D Object Dataset}
Washington RGB-D object dataset includes a total of $41,877$ images for each modality under $51$ object categories and $300$ category instances. Categories are commonly used household objects such as cups, camera, keyboards, vegetables, fruits, etc. Each instance of a category has images taken from $30^\circ$, $45^\circ$ and $60^\circ$ elevation angles. The dataset provides $10$ train/test splits where in each split, one instance for each category is used for testing and the remaining instances are for training. Thus, for a single split run, a total of $51$ category instances (roughly $7,000$ images) are used at testing and the remaining $249$ instances (roughly $35,000$ images) are used at training phase. We evaluate the proposed work on the provided cropped images with the same setup in \cite{Lai_ICRA_2011} for the 10 splits and average accuracy results are reported for the comparison to the related works.
\subsubsection{SUN RGB-D Scene Dataset}
SUN RGB-D scene dataset is the largest real-world RGB-D scene understanding benchmark to the date and  contains RGB-D images of indoor scenes. Following the publicly available configuration of the dataset, we choose $19$ scene categories with a total of $4,845$ images for training and $4,659$ images for testing. We use the same train/test split of \cite{Song_CVPR_2015} to evaluate the proposed work for scene recognition.

\subsection{Object Recognition Performance} \label{sec:exp.objectRecognition}
Table \ref{table:wrgbdResults} shows average accuracy performance of our approach along with the state-of-the-art methods for object recognition on Washington RGB-D object benchmark. Our approach greatly improves the previous state-of-the-art results for both of RGB and depth modalities with a margin of $2.4\%$ and $1.3\%$, respectively. As for the combined RGB-D results, our approach surpasses all the other methods except that of \cite{Loghmani_RAL_2019}, which is slightly better than ours ($0.3\%$). As stated before (see Sec. \ref{sec:relatedwork}), their approach is based on a gated recurrent unit with a set of memory neurons and is powered by a multimodal fusion learning schema. On the other hand, in this paper, we focus on a simple yet effective multi-modal feature extraction framework with a soft voting SVM classification. These results emphasize the importance of deep features in a unified framework based on the incorporation of CNNs and random RNNs. What is interesting here is that even a simple model like AlexNet can yield quite successful results. Concretely, our previous work \citep{Caglayan_ECCVW_2018} with AlexNet architecture, called VGG\_f in MatConvNet toolbox, gives impressive results as the models used in this work.
\begin{table}[!h]
	\caption{Average accuracy comparison of our approach with the related methods on Washington RGB-D Object dataset (\%). \textcolor{red}{Red:} Best result, \textcolor{blue}{Blue:} Second best result, \textcolor{darkgreen}{Green:} Third best result.}
	\begin{center}
		\setlength{\tabcolsep}{0.9em} 
		\def\arraystretch{1.1}
		\begin{adjustbox}{width=\columnwidth}
			\begin{tabular}{ lccc }
				\hline
				Method 											& RGB 							& Depth 						& RGB-D \\ \hline \hline
				Kernel SVM \citep{Lai_ICRA_2011}    				& 74.5 $\pm\hfil$ 3.1 			& 64.7 $\pm\hfil$ 2.2 	 		& 83.9 $\pm\hfil$ 3.5			\\ 
				KDES \citep{Bo_IROS_2011}         				& 77.7 $\pm\hfil$ 1.9 			& 78.8 $\pm\hfil$ 2.7 			& 86.2 $\pm\hfil$ 2.1			\\ 
				CNN-RNN \citep{Socher_NIPS_2012}    				& 80.8 $\pm\hfil$ 4.2 			& 78.9 $\pm\hfil$ 3.8			& 86.8 $\pm\hfil$ 3.3			\\ 
				CaRFs \citep{Asif_ICRA_2015}         			& - 			                & - 			                & 88.1 $\pm\hfil$ 2.4			\\ 
				MMDL \citep{Wang_2015_IEEE_ToM}         			& 74.6 $\pm\hfil$ 2.9			& 75.5 $\pm\hfil$ 2.7			& 86.9 $\pm\hfil$ 2.6			\\ 
				Subset-RNN \citep{Bai_Neurocomp_2015}  			& 82.8 $\pm\hfil$ 3.4 			& 81.8 $\pm\hfil$ 2.6 	 		& 88.5 $\pm\hfil$ 3.1			\\ 
				CNN Features \citep{Schwarz_ICRA_2015}  	        & 83.1 $\pm\hfil$ 2.0 			& -								& 89.4 $\pm\hfil$ 1.3			\\ 
				CNN-SPM-RNN \citep{Cheng_CVIU_2015}        		& 85.2 $\pm\hfil$ 1.2		 	& 83.6 $\pm\hfil$ 2.3 			& 90.7 $\pm\hfil$ 1.1			\\ 
				CFK \citep{Cheng_3DV_2015}  						& 86.8 $\pm\hfil$ 2.7 			& 85.8 $\pm\hfil$ 2.3	        & 91.2 $\pm\hfil$ 1.4			\\ 
				Fus-CNN \citep{Eitel_IROS_2015}  	        	& 84.1 $\pm\hfil$ 2.7 			& 83.8 $\pm\hfil$ 2.7			& 91.3 $\pm\hfil$ 1.4			\\ 
				AlexNet-RNN \citep{Bui_Access_2016}  			& 89.7 $\pm\hfil$ 1.7 			& -								& -								\\ 
				Fusion 2D/3D CNNs \citep{Zia_ICCVW_2017}         & 89.0 $\pm\hfil$ 2.1 			& 78.4 $\pm\hfil$ 2.4			& 91.8 $\pm\hfil$ 0.9			\\ 
				STEM-CaRFs \citep{Asif_ToR_2017}  			    & 88.8 $\pm\hfil$ 2.0 			& 80.8 $\pm\hfil$ 2.1			& 92.2 $\pm\hfil$ 1.3			\\
				MM-LRF-ELM \citep{Liu_Neurocomp_2018}        	& 84.3 $\pm\hfil$ 3.2		 	& 82.9 $\pm\hfil$ 2.5 			& 89.6 $\pm\hfil$ 2.5			\\ 
				VGG\_f-RNN \citep{Caglayan_ECCVW_2018}     		& \bftab\textcolor{darkgreen}{89.9 $\pm\hfil$ 1.6} 	        & 84.0 $\pm\hfil$ 1.8			& 92.5 $\pm\hfil$ 1.2	        \\
				DECO \citep{Carlucci_RAS_2018}     		        & 89.5 $\pm\hfil$ 1.6 	        & 84.0 $\pm\hfil$ 2.3			& \bftab\textcolor{darkgreen}{93.6 $\pm\hfil$ 0.9}	        \\
				MDSI-CNN \citep{Asif_TPAMI_2018}  			    & \bftab\textcolor{darkgreen}{89.9 $\pm\hfil$ 1.8} 			& 84.9 $\pm\hfil$ 1.7			& 92.8 $\pm\hfil$ 1.2			\\
				HP-CNN \citep{Zaki_AuotRobots_2019}    		    & 87.6 $\pm\hfil$ 2.2 			& 85.0 $\pm\hfil$ 2.1			& 91.1 $\pm\hfil$ 1.4			\\ 
				RCFusion \citep{Loghmani_RAL_2019}  			& 89.6 $\pm\hfil$ 2.2 			& \bftab\textcolor{darkgreen}{85.9 $\pm\hfil$ 2.7}			& \bftab\textcolor{red}{94.4 $\pm\hfil$ 1.4}			\\
				MMFLAN \citep{Qiao_2021_Neuroc}  				& 83.9 $\pm\hfil$ 2.2 			& 84.0 $\pm\hfil$ 2.6			& 93.1 $\pm\hfil$ 1.3			\\ \hline
				\bftab{This work - AlexNet-RNN}     			& 83.0 $\pm\hfil$ 1.9 	        & 84.1 $\pm\hfil$ 2.3			& 90.9 $\pm\hfil$ 1.3	            \\
				\bftab{This work - DenseNet121-RNN}     		& \bftab\textcolor{blue}{91.5 $\pm\hfil$ 1.1} 	        & \bftab\textcolor{blue}{86.9 $\pm\hfil$ 2.1}			& 93.5 $\pm\hfil$ 1.0	            \\
				\bftab{This work - ResNet101-RNN}     			& \bftab\textcolor{red}{92.3 $\pm\hfil$ 1.0} 	       & \bftab\textcolor{red}{87.2 $\pm\hfil$ 2.5}			& \bftab\textcolor{blue}{94.1 $\pm\hfil$ 1.0}	            \\
				\hline
			\end{tabular}
		\end{adjustbox}
		\label{table:wrgbdResults}
	\end{center}
\end{table}

We also present average accuracy performance of individual object categories on the 10 evaluation splits of Washinton RGB-D Object dataset using the best-performing structure, ResNet101-RNN. As shown in Fig. \ref{fig:wrgbdIndividualResults}, our approach is highly accurate in recognition of the most of the object categories. Categories with lower accuray results are \textit{mushroom}, \textit{peach}, and \textit{pitcher}. The common reason that leads to the lower performance in these categories seems to be due to their less number of instances. In particular, these categories have only $3$ instances, which is the minimum number for any category in the dataset. Considering the other categories with up to $14$ instances, this imbalance of the data may have biased the learning to favor of categories with more examples. Moreover, the accuracy of our combined RGB and depth based on weighted confidences of modalities reflects that the fusion of RGB and depth data in this way can provide strong discrimination capability for object categories.

\subsection{Scene Recognition Performance} \label{sec:exp.sceneRecognition}
To test the generalization ability of our approach, we also carry out comparative analysis of our best-performing model, namely ResNet101-RNN, on SUN RGB-D Scene \citep{Song_CVPR_2015} dataset for scene recognition as a more challenging task of scene understanding. To this end, we first apply ResNet101 pretrained model without finetuning, namely Fix ResNet101-RNN, for both of RGB and depth modalities. Then, we finetune the pretrained CNN model on SUN RGB-D Scene dataset using the same hyper-parameters of object recognition task (see Sec. \ref{sec.exp.ma.finetuning}). The results of these experiments together with the-state-of-the-art results on this dataset are reported in Table \ref{table:sunrgbdResults}. Our best system greatly improves the-state-of-the-art methods not only for RGB-D final result but also for individual data modalities. It is worth mentioning that we use the pretrained CNN model on object-centric dataset of ImageNet \citep{Deng_Imagenet_CVPR_2009}, which is less commonly used for scene recognition task than the pretrained models on scene-centric datasets such as Places \citep{Zhou_NIPS_2014}. Nevertheless, our approach produces superior results compared to existing state-of-the-art methods for RGB-D scene recognition task. Moreoever, it is interesting that our system even with fixed pretrained CNN model is already discriminative enough and achieves impressive accuracy performances. Contrary to our findings on Washington RGB-D Object dataset, finetuning provides much better results not only for depth domain but also for the RGB domain as well. This is what we expect as scene recognition is a cross-domain task for our approach that has the pretrained CNN model of the object-centric ImageNet as the backbone. Specifically, finetuning on depth data boosts the accuracy greatly by providing both domain and modality adaptation.
\begin{table}[!h]
	\caption{Accuracy comparison of our approach with the related methods on SUN RGB-D Scene dataset (\%). \textcolor{red}{Red:} Best result, \textcolor{blue}{Blue:} Second best result, \textcolor{darkgreen}{Green:} Third best result.}
	\begin{center}
		\setlength{\tabcolsep}{0.9em} 
		\def\arraystretch{1.2}
		\begin{adjustbox}{width=\columnwidth}
			\begin{tabular}{ lccc }
				\hline
				Method 											& RGB 				& Depth 			& RGB-D \\ \hline \hline
				Places CNN-Lin SVM \citep{Zhou_NIPS_2014}    	& 35.6 				& 25.5 	 			& 37.2 		\\ 
				Places CNN-RBF SVM \citep{Zhou_NIPS_2014}    	& 38.1 				& 27.7 	 			& 39.0 		\\ 
				SS-CNN-R6 \citep{Liao_ICRA_2016}    			& 36.1 				& - 	 			& 41.3 		\\ 
				DMFF \citep{Zhu_CVPR_2016}    					& 37.0  			& - 				& 41.5 		\\ 
				Places CNN-RCNN \citep{Wang_CVPR_2016}         	& 40.4 	           	& 36.3 	        	& 48.1 		\\ 
				MSMM \citep{Song_IJCAI_2017}         			& 41.5 	           	& 40.1 	        	& 52.3 		\\ 
				RGB-D-CNN \citep{Song_AAAI_2017}         		& 42.7 	           	& 42.4 	        	& 52.4 		\\ 
				MDSI-CNN \citep{Asif_TPAMI_2018}         		& 39.6 	           	& 35.2 	        	& 45.2 		\\ 
				DF\textsuperscript{2}Net \citep{Li_AAAI_2018df} & - 	           	& - 	        	& 54.6 		\\ 
				HP-CNN-T \citep{Zaki_AuotRobots_2019}         	& 38.8 	           	& 28.5 	        	& 42.2 		\\ 
				LM-CNN \citep{2019_CogComp_Cai}         		& 44.3 	           	& 34.6 	        	& 48.7 		\\ 
				RGB-D-OB \citep{Song_TIP_2019}         			& - 	           	& 42.4 	        	& 53.8 		\\ 
				Cross-Modal Graph \citep{Yuan_2019_AAAI}        & 45.7 	           	& - 	        	& 55.1		\\
				RAGC \citep{Montoro_2019_ICCV} 					& - 	           	& - 	        	& 42.1 		\\ 
				MAPNet \citep{2019_PR_Li} 						& - 	           	& - 	        	& 56.2 		\\ 
				TRecgNet Aug \citep{Du_2019_CVPR}         		& 50.6 				& \bftab\textcolor{darkgreen}{47.9} 	& 56.7 		\\
				G-L-SOOR \citep{Song_TIP_2020}         			& 50.5 	           	& 44.1 	        	& 55.5		\\
				MSN \citep{2020_Neuroc_Xiong}         			& - 	           	& - 	        	& 56.2		\\
				CBCL \citep{Ayub_2020_BMVC}         			& 48.8 				& 37.3 	        	& \bftab\textcolor{darkgreen}{59.5} 		\\ 
				ASK \citep{2021_TIP_Xiong}  					& - 				& -	        		& 57.3 					\\
				2D-3D FusionNet \citep{2021_IF_Montoro}  		& \bftab\textcolor{blue}{56.4} 			& 44.1	        		& 58.6 					\\
				TRecgNet Aug - ResNet18 \citep{Du_2021_IJCV}  	& 53.8 				& \bftab\textcolor{blue}{49.3}	        	& 58.5 					\\
				TRecgNet Aug - ResNet101 \citep{Du_2021_IJCV}  	& \bftab\textcolor{darkgreen}{54.2} 			& \bftab\textcolor{blue}{49.3} 	        	& \bftab\textcolor{blue}{59.8} 					\\ \hline
				\bftab{This work - Fix ResNet101-RNN}           & 50.8 				& 38.6 				& 53.1 		\\ 
				\bftab{This work - Finetuned ResNet101-RNN}  	& \bftab\textcolor{red}{58.5}  			& \bftab\textcolor{red}{50.1}  			& \bftab\textcolor{red}{60.7} 	\\ 
				\hline
			\end{tabular}
		\end{adjustbox}
		\label{table:sunrgbdResults}
	\end{center}
\end{table}

\begin{figure}[!ht]
	\begin{center}
		\includegraphics[width=\columnwidth, keepaspectratio]{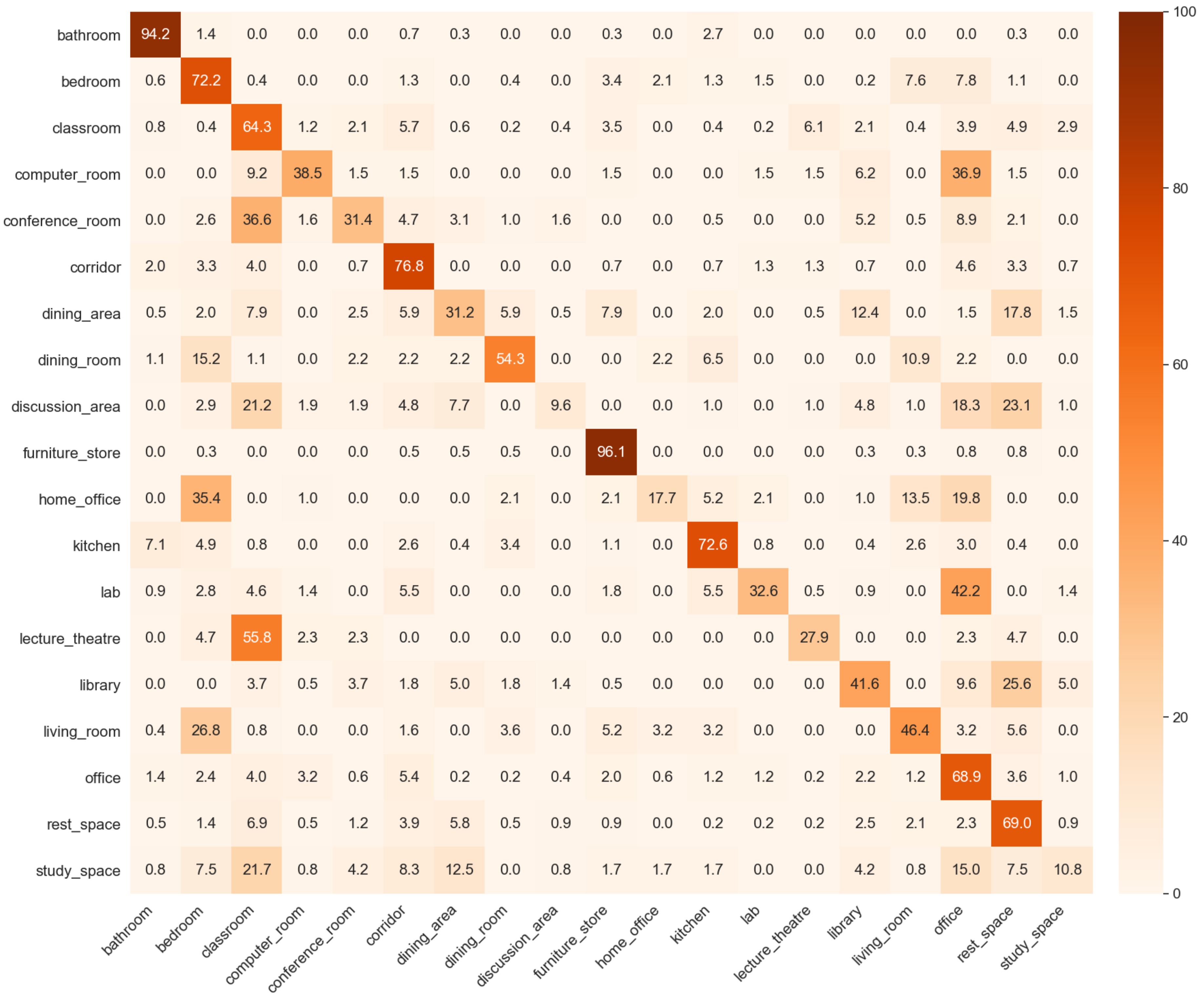}
	\end{center}
	\caption{RGB-D confusion matrix of ResNet101-RNN on SUN RGB-D Scene dataset (best viewed with magnification).}
	\label{fig:sunrgbdConfusionMatrix}
\end{figure}

\begin{figure}[!t]
	\begin{center}
		\includegraphics[width=0.95\columnwidth, keepaspectratio]{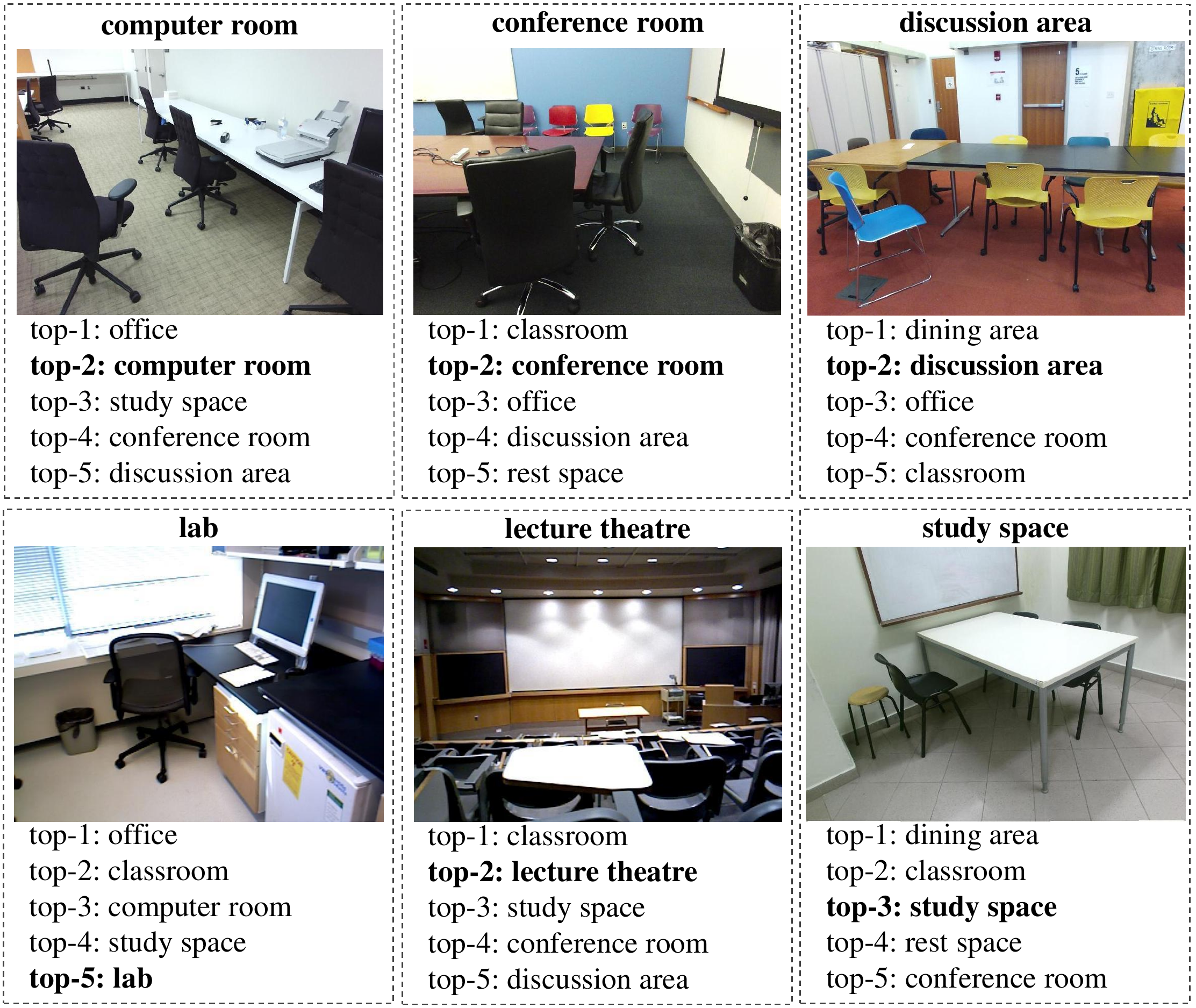}
	\end{center}
c	\caption{Top-5 RGB-D predictions of our system using sample test images of frequently confused scene categories on SUN RGB-D Scene dataset.}
	\label{fig:sunrgbdConfusedSamples}
\end{figure}

Fig. \ref{fig:sunrgbdConfusionMatrix} shows the confusion matrix of our approach with finetuning over the $19$ categories of SUN RGB-D Scene dataset for RGB-D. The matrix demonstrates the degree of confusion between pairs of scene categories and implies the similarity between scenes on this dataset. The largest misclassification errors happen to be between extremely similar scene categories such as \textit{computer room} - \textit{office}, \textit{conference room}-\textit{classroom}, \textit{discussion area}-\textit{rest space}, \textit{lecture theatre}-\textit{classroom}, \textit{study space}-\textit{classroom}, \textit{lab}-\textit{office}, etc. In addition to the inter-class similarity, other reasons for poor performance might be intra-class variations of the scenes and lack of getting enough representative knowledge transfer from the ImageNet models.
To further analyse the performance of our system, we give top-3 and top-5 classification accuracy together with top-1 results as in Table \ref{table:top135sceneResults}. While the top-1 accuracy shows the percentage of test images that exactly matches with the predicted classes, the top-3 and top-5 indicates the percentage of test images that are among the top ranked 3 and 5 predictions, respectively. The top-3 and top-5 results demonstrate the effectiveness of our system more closely by overcoming ambiguity among scene categories greatly. Fig. \ref{fig:sunrgbdConfusedSamples} depicts some test examples of scene categories confused with each other frequently on SUN RGB-D Scene dataset. As shown in the figure, these scene categories have similar appearances that make them hard to distinguish even for a human expert without sufficient context knowledge in the evaluation. Nevertheless, our approach is able to identify scene category labels among the top-3 and top-5 predictions with high accuracy.
\begin{table}[!h]
	\caption{Scene recognition accuracy of top-1, top-3, and top-5 on SUN RGB-D Scene dataset (\%).}
	\begin{center}
		\setlength{\tabcolsep}{0.9em} 
		\def\arraystretch{1.2}
		\begin{adjustbox}{width=0.65\columnwidth}
			\begin{tabular}{ lccc }
				\hline
				Accuracy			& RGB 				& Depth 			& RGB-D \\ \hline \hline
				top-1    			& 58.5 				& 50.1 	 			& 60.7 		\\ 
				top-3    			& 81.0 				& 71.5 	 			& 83.5 		\\ 
				top-5    			& 88.5 				& 80.9 	 			& 89.9 		\\ 
				\hline
			\end{tabular}
		\end{adjustbox}
		\label{table:top135sceneResults}
	\end{center}
\end{table}
\subsection{Model Ablation} \label{sec:exp.modelAblation}
\begin{figure*}[!ht]
	\centering
	\subfloat{\includegraphics[width=\columnwidth, keepaspectratio]{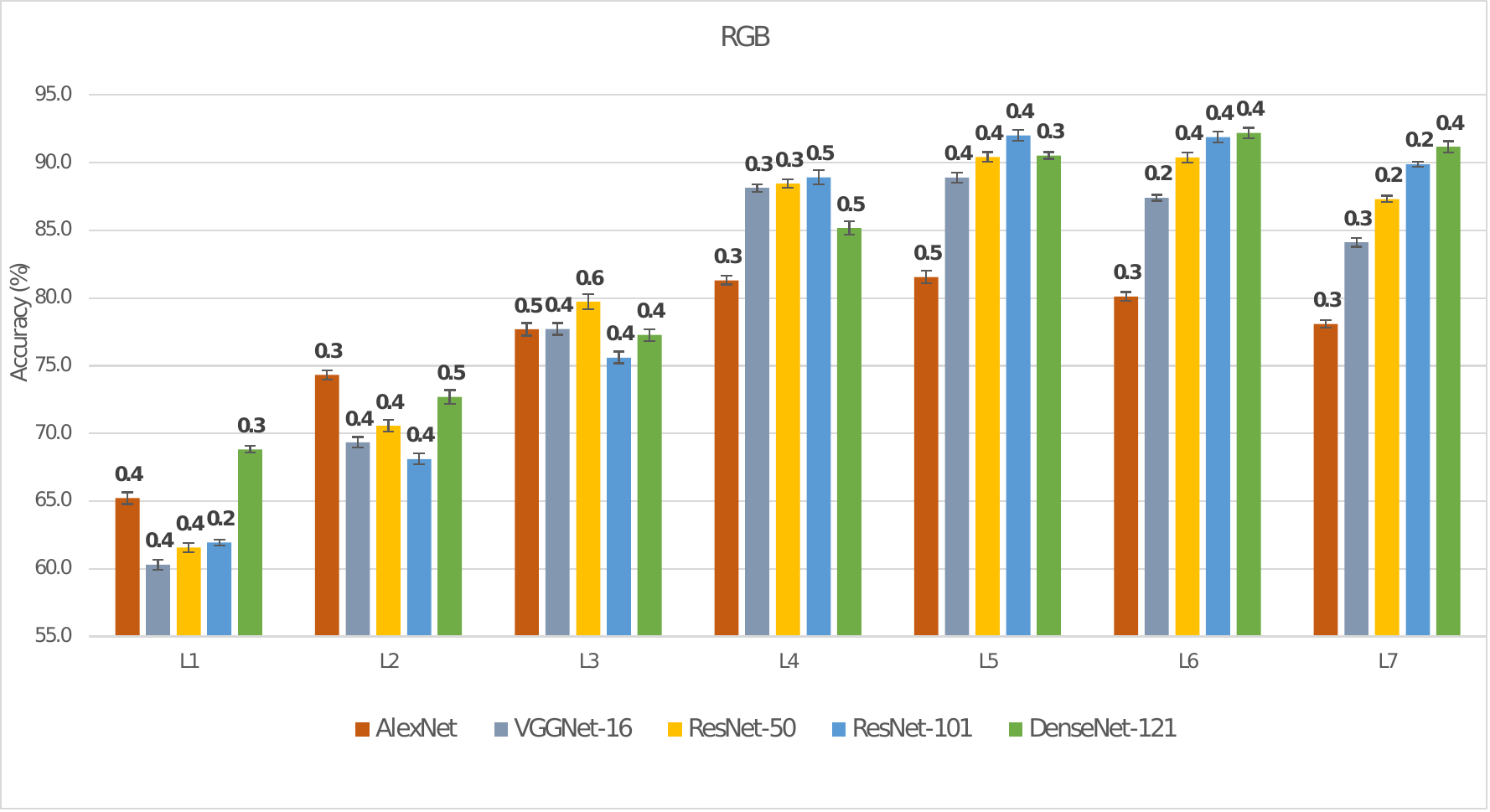}}%
	\subfloat{\includegraphics[width=\columnwidth, keepaspectratio]{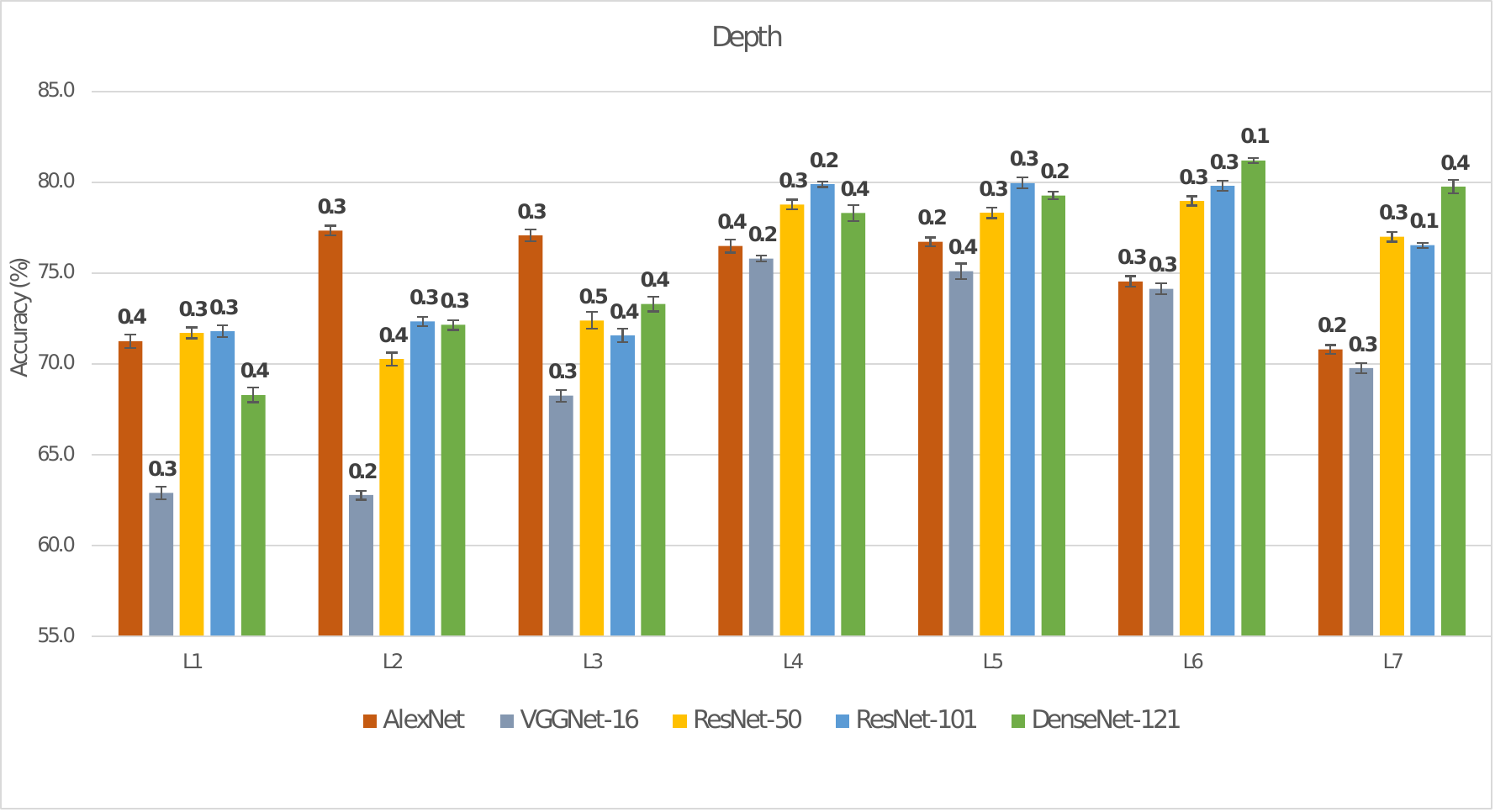}}%
	\caption{Effect of randomness on the accuracy results for each level (L1 to L7). Values indicate standard deviations.}
	\label{fig:randomness}
\end{figure*}
We have analyzed and validated the proposed framework with extensive experiments using a variety of architectural configurations on the popular benchmark of Washington RGB-D dataset, which is more than $4$ times larger than the SUN RGB-D scene dataset. In this section, the analysis and evaluations of the model ablative investigations are presented. Further experiments and analysis are given in the \textit{supplementary material}. The developmental experiments are carried out on two splits of Washington RGB-D Object dataset for both modalities in order to evaluate on more stable results. The average results are analyzed. However, in some experiments, more runs have been carried out, which are clearly stated in the related sections. In Sec. \ref{sec:exp.objectRecognition} and Sec. \ref{sec:exp.sceneRecognition}, the best performing models are compared with the state-of-the-art methods with the exact provided evaluation setups. We assess the proposed framework on a PC with AMD Ryzen 9 3900X 12-Core Processor, 3.8 GHz Base, 128 GB DDR4 RAM 2666 MHz, and NVIDIA GeForce GTX 1080 Ti graphics card with 11 GB memory.
\begin{figure*}[!b]
	\centering
	 \subfloat{\includegraphics[width=\columnwidth, keepaspectratio]{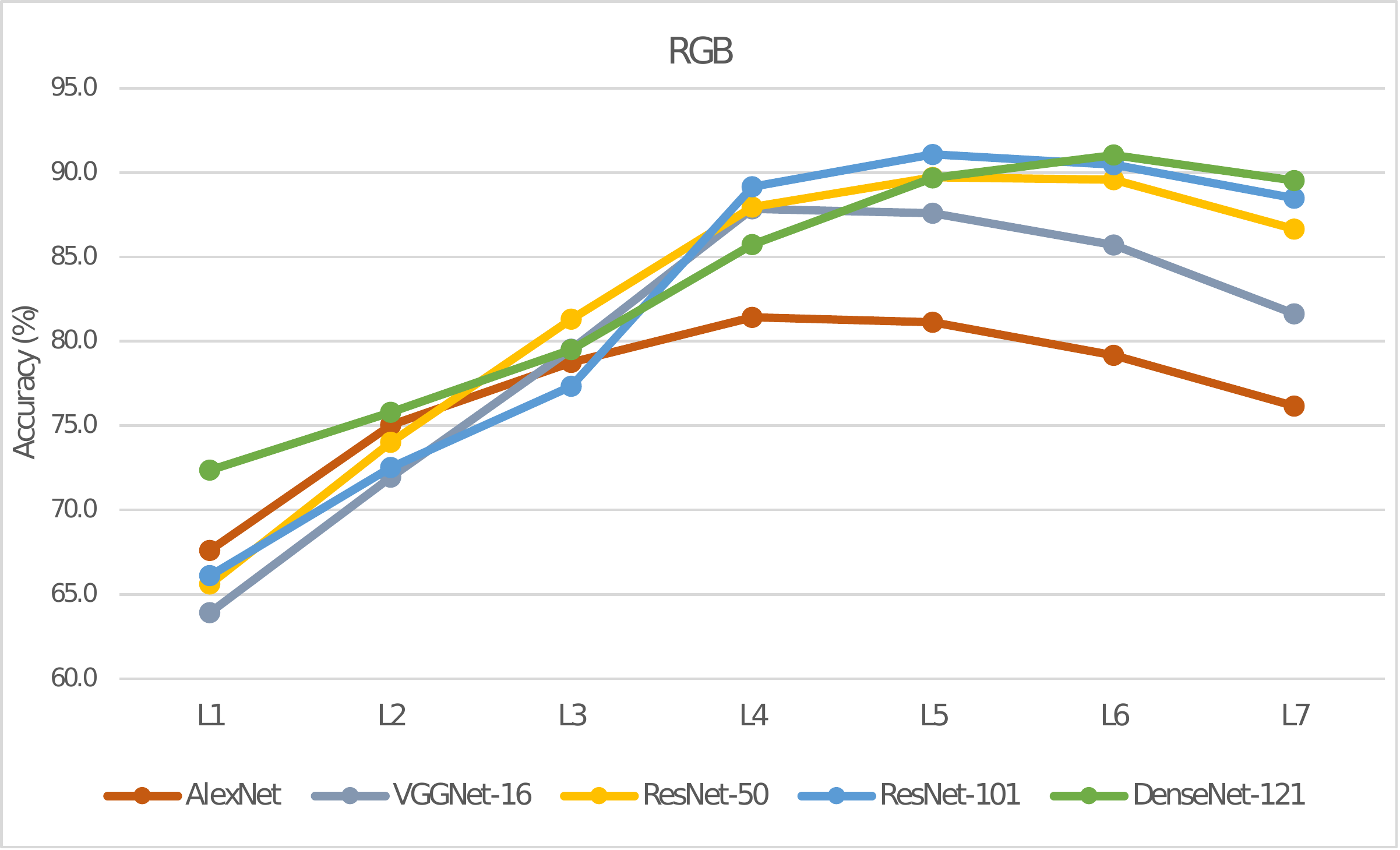}}%
   \subfloat{\includegraphics[width=\columnwidth, keepaspectratio]{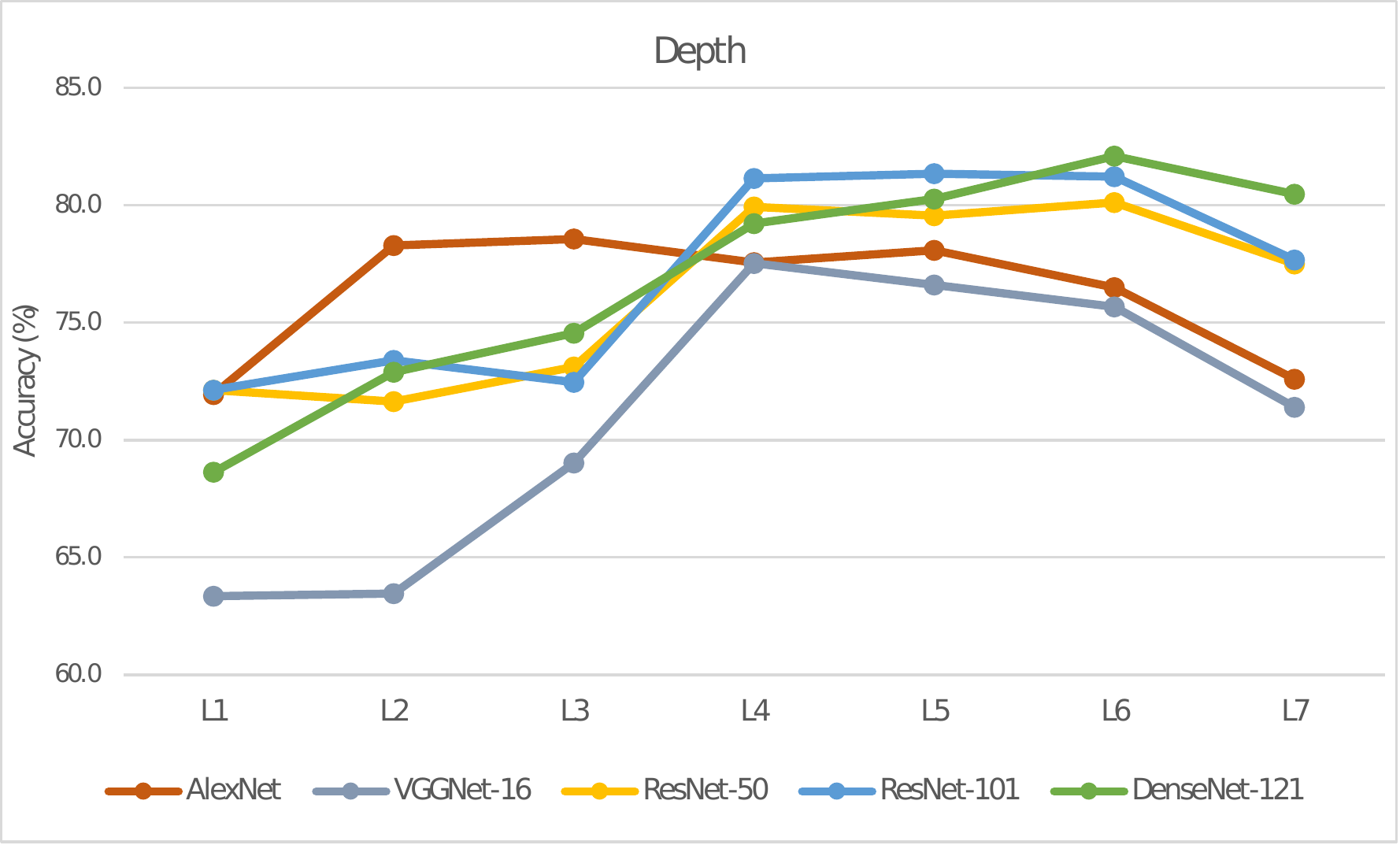}}%
l	\caption{Level-wise average accuracy performance of different baseline models on all the 10-splits of Washington RGB-D dataset.}
	\label{fig:levelwisePerformances}
\end{figure*}
\subsubsection{Empirical Evaluation of the Effect of Randomness}
\label{sec.exp.ma.randomness}
The use of random weights both in pooling and RNN structures leads to the question of how stable are the results. Thus, we experimentally investigate to see whether there is a decisive difference between different runs that generate and use new random weights. We run the pipeline with different random weights on two splits, 5 times for each. Fig. \ref{fig:randomness} reports average results with their standard deviations for each level. The figure clearly shows that randomness does not cause any instability in the model and produces similar results with very small deviations.

\subsubsection{Level-wise Performance of Different Models} \label{sec.exp.ma.levelPerformances}
Fig. \ref{fig:levelwisePerformances} shows level-wise average accuracy performances of all the baseline models for both of RGB and depth modalities on all the 10 evaluation splits. The graphs show a similar performance trend line with an upward at the beginning and a downward at the end. Although the levels at which optimum performance is obtained vary according to the model, what is common to all models in general is that instead of final level representations, intermediate level representations present the optimal results. These experiments also verify that while deep models transform attributes from general to specific through  the network eventually \citep{Razavian_CVPRW_2014, Zeiler_ECCV_2014}, intermediate layers present the optimal representations. This makes sense because while early layers response to low-level raw features such as corners and edges, late layers extract more object-specific features of the trained datasets. This is more clear on the depth plot in Fig. \ref{fig:levelwisePerformances}, where the dataset difference is obvious due to the domain difference. We should state that RNN encoding on features extracted from FC layers with less than $8192$ dimension might not be efficient since they are already compact enough. Therefore, encoding outputs of these layers to a larger feature space through RNNs might lead to redundancy in representations. This might be another reason why there is a drop in accuracy of these layers (e.g. see L7 in Fig. \ref{fig:levelwisePerformances}). In addition, depth plot contains more fluctuations and irregularities comparing to the RGB plot, since the pretrained models of the RGB ImageNet are used as fixed extractors without finetuning. As for the different baseline model comparison, ResNet-101 and DenseNet-121 models perform similarly in terms of accuracy and are better than others.
\begin{figure*}[!hb]
	\centering
	 \subfloat{\includegraphics[width=\columnwidth, keepaspectratio]{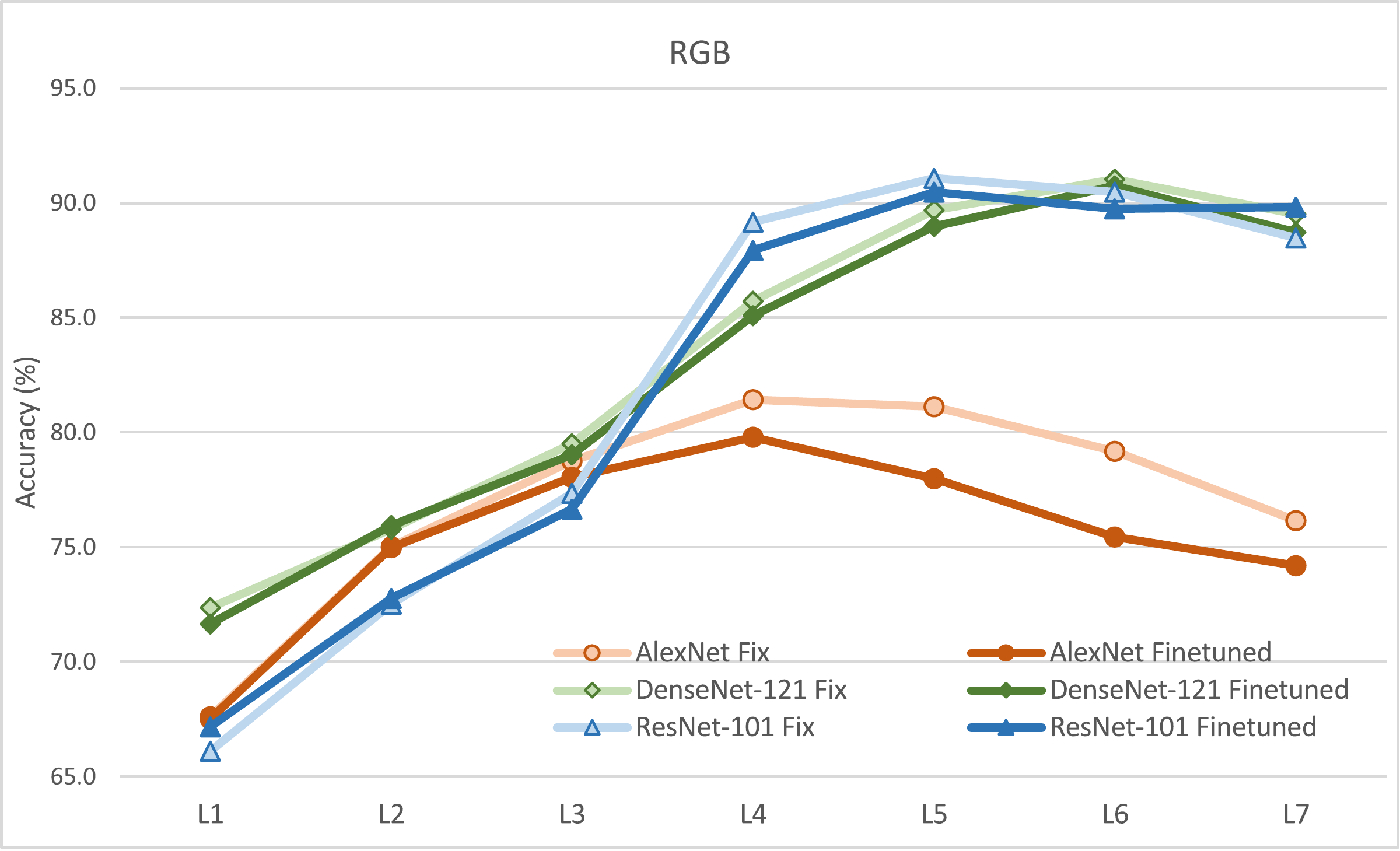}}%
	\subfloat{\includegraphics[width=\columnwidth, keepaspectratio]{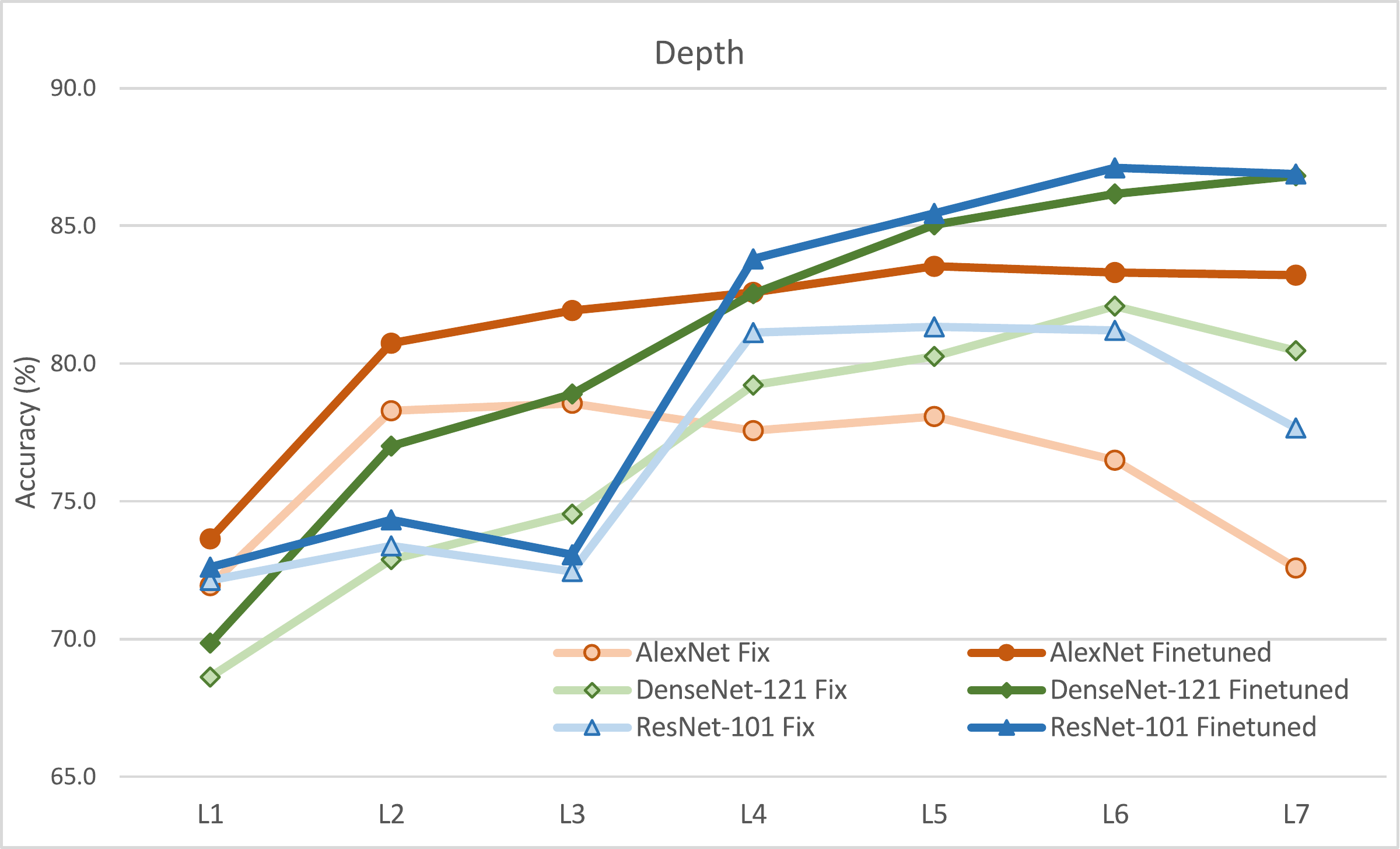}}%
f	\caption{Level-wise average accuracy performance of finetuned CNN models together with fixed models on all the 10-splits of Washington RGB-D dataset.}
	\label{fig:finetuning}
\end{figure*}
\subsubsection{Comparative Results of Random Weighted Pooling} \label{sec.exp.ma.poolingPerformances}
\begin{figure}[!ht]
	\centering
	\includegraphics[width=\columnwidth, keepaspectratio]{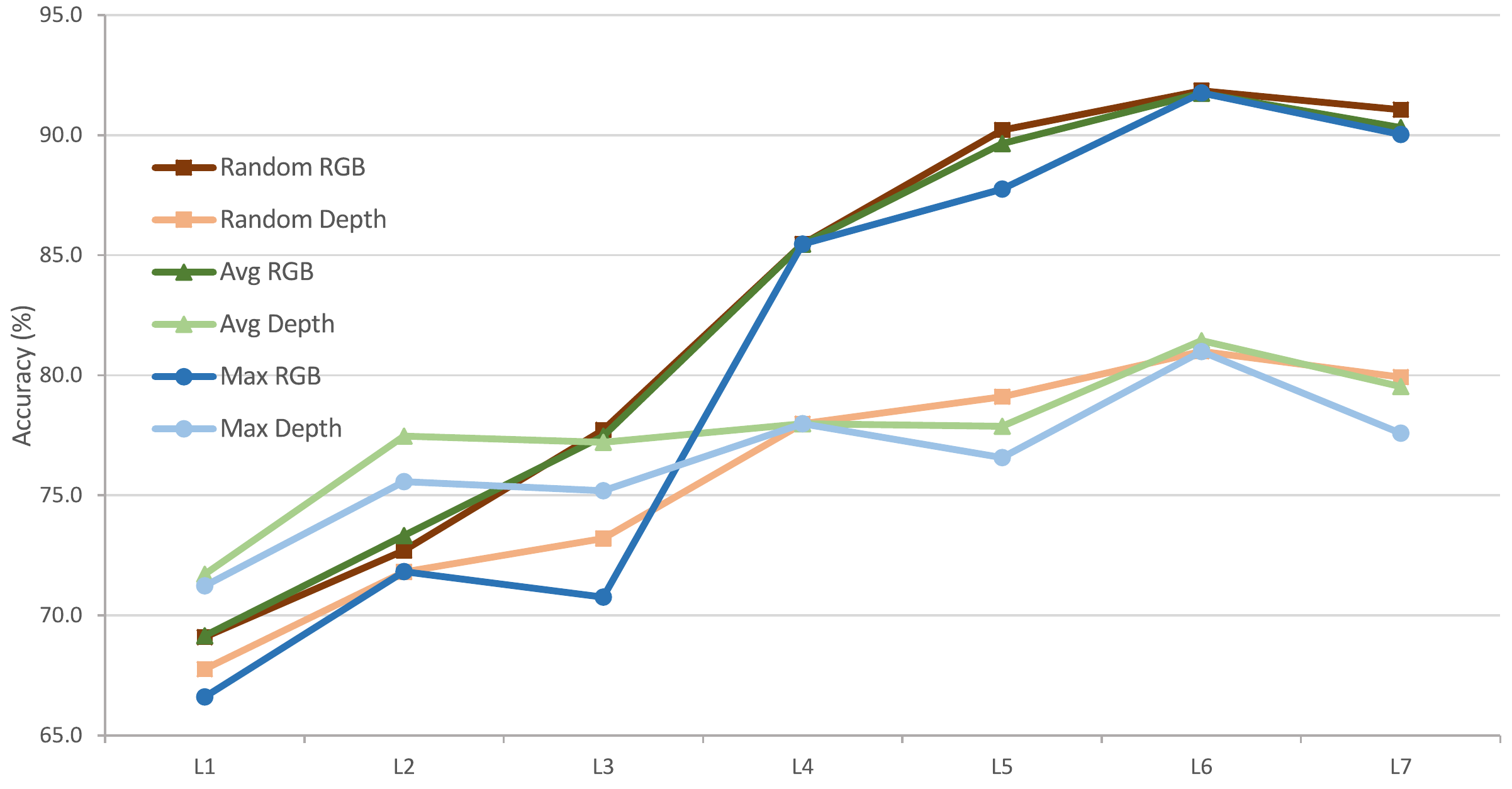}
p	\caption{Average accuracy performance of different pooling methods on RGB and depth data for the baseline model of DenseNet-121 on two splits of Washington RGB-D dataset.}
	\label{fig:poolingComparison}
\end{figure}

In our approach, we extend the idea of randomness into a pooling strategy to cope with the high dimensionality of CNN activations, which could not be only applied to map/window size but also can be used to reduce the number of maps. We particularly employ random pooling to confirm that randomness works greatly in overall RNN-Stage even in such a pooling strategy together with random RNNs. To this end, we investigate the comparative accuracy performances of random pooling together with average pooling and max pooling. We use the DenseNet-121 model, where pooling is used extensively on each level (except in level 4), and we conduct experiments using the same RNN weights for fair comparison. Fig. \ref{fig:poolingComparison} shows average accuracy results of two splits for each pooling on both RGB and depth data. As seen from the figure, random weighted pooling generally performs similar to average pooling, while its performance in average is better than max pooling. Moreover, it is seen that random pooling acquires better results especially in middle/late levels(L4-L7), which presents more stable and semantically meaningful representations comparing to the early levels. The results also show that the proposed random pooling and average pooling can be used interchangeably as their performances are similar.

We further investigate the comparative accuracy performance of the proposed random pooling in our final ResNet-101 based pipeline. As it can be seen in Table~\ref{table:poolingComparison}, when the proposed pipeline is set to use random weighted pooling, it produces better or similar accuracy than max pooling and average pooling-based pipelines. This validates the power of randomness in a pooling strategy and the use of random pooling as an alternative way for down-sampling.
\begin{table}[!h]
	\caption{Average accuracy performance of different pooling methods in the best performing ResNet101-RNN pipeline on Washington RGB-D dataset (\%).}
	\begin{center}
		\setlength{\tabcolsep}{0.9em} 
		\def\arraystretch{1.2}
		\begin{adjustbox}{width=0.75\columnwidth}
			\begin{tabular}{ lccc }
				\hline
				Accuracy			& RGB 								& Depth 							& RGB-D \\ \hline \hline
				Max   				& 91.1 $\pm\hfil$ 1.4 				& 87.1 $\pm\hfil$ 2.5 	 			& 93.8 $\pm\hfil$ 0.9 		\\ 
				Average    			& 91.6 $\pm\hfil$ 1.6 				& 87.2 $\pm\hfil$ 2.5 	 			& 94.0 $\pm\hfil$ 1.0 		\\ 
				Random   			& 92.3 $\pm\hfil$ 1.0 				& 87.2 $\pm\hfil$ 2.5 	 			& 94.1 $\pm\hfil$ 1.0 		\\ 
				\hline
			\end{tabular}
		\end{adjustbox}
		\label{table:poolingComparison}
	\end{center}
\end{table}

\subsubsection{Contribution of Finetuning} \label{sec.exp.ma.finetuning}
We have not used any training or finetuning in our approach to feature extraction in the ablative experiments so far (except Table~\ref{table:poolingComparison}, where depth modality-based ResNet101 is finetuned). Although impressive results are obtained on RGB data, the same success is not achieved on depth data. The reason for this difference is that the baseline CNN models are pretrained models on RGB dataset of the ImageNet. Therefore, as the next step, we analyze the changes in accuracy performance of RGB and depth data modalities by finetuning the baseline CNN models in our approach. To this end, we first carry out a systematic inquiry to find optimal finetuning hyper-parameters on a predefined set of values using only one split of Washington RGB-D dataset as a validation set for AlexNet and DenseNet-121 models. Then, finetuning of the models are performed by stochastic gradient descent (SGD) with momentum. The hyper-parameters of momentum, learning rate, batch size, learning rate decay factor and decay step size, and number of epochs, respectively are used as following; $(0.9, 0.001, 32, 0.01, 10, 40)$ and $(0.9, 0.0001, 8, 0.1, 10, 40)$ are used for AlexNet on RGB and depth data, respectively, whereas $(0.95, 0.0001, 16, 0.1, 10, 40)$ and $(0.95, 0.001, 8, 0.1, 10, 40)$ are used for DenseNet-121. Apart from these two models, we also perform finetuning on the ResNet-101 model. We use the same finetuning hyperparameters of DenseNet-121 for ResNet-101, since they are in a similar architectural structure. Fig. \ref{fig:finetuning} shows average accuracy performance of finetuned CNN models together with fixed models on all the 10 evaluation splits of Washington RGB-D object dataset. The plot shows a clear upward in performance on depth data as expected. However, there is a loss of accuracy in general, when finetuning is performed on RGB data. Washington RGB-D object dataset contains a subset of the categories in ImageNet. Accordingly, pretrained models of ImageNet already satisfy highly correlated distribution on RGB data. Therefore, there is no need for finetuning on RGB data. In contrast, in order to ensure coherence and relevance, finetuning is required for depth data due to domain difference of the inputs with the pretrained models.

\subsubsection{Weighted Voting based RGB-D Fusion Performance} \label{sec.exp.ma.weightedFusion}
Finally, we provide RGB-D combined results for AlexNet, DenseNet-121, and ResNet-101 models as shown in Table \ref{table:rgbdFusions} based on the SVM confidences. The table reports average results for fusion of the best levels of RGB and depth, and the best trio levels (see the \textit{supplementary material}). We evaluate two types of soft voting, our proposed weighted vote and average vote. The proposed weighted vote increases accuracy comparing to the average vote for all the models both on the multi-modal fusion of the best single and best trio levels of RGB and depth streams. The results also confirm the strength of our multi-modal voting approach that combines RGB and depth modalities effectively. On the other hand, the reason why RGB-D fusion improves the individual RGB and depth results lies in the fact that these different data modalities support each other towards a more accurate representation by capturing different aspects of the data with a strong complementary approach. RGB data are rich in terms of texture and color information. Depth data have additional geometric information to depict object shapes. Moreover, depth sensors are more insensitive to changes in lighting conditions. Therefore, multi-modal data combination is useful not only for its integrative characteristic, but also for its complementarity when one modality data is lacking such as RGB data in dark environment or depth data on shiny surfaces.
\begin{table}[!h]
	\caption{Average accuracy performance of RGB-D (RGB + Depth) with different fusion combinations on Washington RGB-D dataset (\%).}  
	\label{table:rgbdFusions}    
	\centering
	\setlength{\tabcolsep}{0.9em} 
	\def\arraystretch{1.2}
	\begin{adjustbox}{width=\columnwidth}
		\begin{tabular}{llccc}
			\hline
			&								       																				& AlexNet 						  & DenseNet-121 					& ResNet-101 				\\ \hline \hline
			Avg Vote 		& $\scalebox{.9}{RGB}_{\scalebox{.7}{LB1}} \kern 3.3em+ \scalebox{.9}{Depth}_{\scalebox{.7}{LB1}}$          		& 90.2 $\pm\hfil$ 1.3        	  & 92.9 $\pm\hfil$ 1.4           	& 92.7 $\pm\hfil$ 1.6         \\
			Weighted Vote 	& $\scalebox{.9}{RGB}_{\scalebox{.7}{LB1}} \kern 3.3em+ \scalebox{.9}{Depth}_{\scalebox{.7}{LB1}}$     			   	& 90.2 $\pm\hfil$ 1.2        	  & 93.5 $\pm\hfil$ 1.0          	& 93.8 $\pm\hfil$ 1.1          \\
			Avg Vote 		& $\scalebox{.9}{RGB}_{\scalebox{.7}{LB1+LB2+LB3}} \kern 0.2em+ \scalebox{.9}{Depth}_{\scalebox{.7}{LB1+LB2+LB3}}$  & 90.6 $\pm\hfil$ 1.6        	  & 92.6 $\pm\hfil$ 1.4          	& 93.0 $\pm\hfil$ 1.3           \\
			Weighted Vote 	& $\scalebox{.9}{RGB}_{\scalebox{.7}{LB1+LB2+LB3}} \kern 0.2em+ \scalebox{.9}{Depth}_{\scalebox{.7}{LB1+LB2+LB3}}$  & \textbf{90.9 $\pm\hfil$ 1.3}    & \textbf{93.5 $\pm\hfil$ 1.0}    & \textbf{94.1 $\pm\hfil$ 1.0} 	 \\ \hline
			\multicolumn{5}{l}{LB1: Best performing level \quad LB2: Second best performing level \quad LB3: Third best performing level}
		\end{tabular}
	\end{adjustbox}
\end{table}

\subsection{Discussion} \label{sec:exp.discussion}
Our framework presents an effective solution for deep feature extraction in an efficient way by integrating a pretrained CNN model with random weights based RNNs. Randomization throughout our RNN-Stage raises the question of whether the results are stable enough. The carefully implemented experiments in Sec. \ref{sec.exp.ma.randomness} are an empirical justification for the stability of random weights. On the other hand, our multi-level analysis shows that the optimum performance gain from a single level always comes from an intermediate level for all the models with/without finetuning for both of RGB and depth modalities. The only exception is in the use of finetuned DenseNet-121 model on depth data. This is an interesting finding, because one expects better representation capabilities of final layers, especially in the use of finetuned models. Yet, as expected, performance generally increases from the first level to the last level throughout the networks when the underlying CNN models are finetuned. Since Washington RGB-D Object \citep{Lai_ICRA_2011} dataset includes a subset of object categories in the ImageNet \citep{Deng_Imagenet_CVPR_2009}, finetuning does not improve accuracy success on RGB data. In contrast, accuracy gain is significant due to the need for domain adaptation in depth data. This also shows that using an appropriate technique to handle depth data as in our approach (see the \textit{supplementary material}), leads impressive performance improvement by knowledge transfer between modalities. 

In this study, although we have explored different techniques to fuse representations of multiple levels to further increase the classification success, a single optimum level may actually be sufficient enough for many tasks. In this way, especially for tasks where computational time is more critical, results can be obtained much faster without sacrificing accuracy success. Another point of interest is that the data imbalance in Washington RGB-D Object dataset results in poor performance for the individual categories with less instances and consequently leads to a drop in the overall success of the system. Hence, this imbalance might be overcome by applying data augmentation on the categories with less instances.

The success of our approach for RGB-D scene recognition confirms the generalization ability of the proposed framework. Unlike object recognition, when the underlying CNN models are finetuned, success in both RGB and depth modalities increases significantly in scene recognition task. This is due to the need for cross-domain task adaptation of object-centric based pretrained models. Therefore, similar findings in object recognition could be observed if scene-centric based pretrained models are employed for scene recognition (e. g. Places \citep{Zhou_NIPS_2014}). Moreover, such pretrained models could improve the results further within our framework. Another potential that could improve the success for scene recognition is embedding contextual knowledge by jointly employing attention mechanism such as \cite{Fukui_2019_CVPR} in our structure.

This work has been implemented as the extension of our previous work \citep{Caglayan_ECCVW_2018}. Therefore, we have not explored further multimodal architectures. However, instead of SVM, combining level-wise outputs through a multilayer perceptron (MLP) might be more convenient for RGB-D multimodal design. In particular, it would be interesting to use the soft voting approach proposed in this study with MLP. In the future, we plan to investigate such an approach for a better RGB-D multimodal tasks, such that success is focused on the ultimate RGB-D fusion rather than the individual accuracy success of the RGB and depth modalities.

\section{Conclusion}
\label{sec:conclusion}
In this paper, we have presented a framework that incorporates pretrained CNN models together with multiple random recursive neural networks. The proposed approach greatly improves RGB-D object and scene recognition performances over the-state-of-the-art results in the literature on the widely used Washington RGB-D Object and SUN RGB-D Scene datasets. The proposed randomized pooling schema allows us to deal with high-dimensional activations of CNN models effectively. The extensive experimental analysis of various parameters and setup properties show that the incorporation of multiple random RNNs with a pretrained CNN model provides a robust and effective general solution for both of RGB-D object and scene recognition tasks. Utilizing depth data by mapping it into RGB-like image domain allows knowledge transfer from RGB pretrained CNN models effectively. The generic design and the generalization capability of the proposed framework allow to utilize it for other visual recognition tasks. Thus, we have opened our code along with models to the community in order to help future studies.

\section*{Acknowledgment}

This paper is based on the results obtained from a project commissioned by the New Energy and Industrial Technology Development Organization (NEDO).




\bibliographystyle{model2-names}
\bibliography{cviu}

\end{document}


%
\title{\textbf{Supplementary Material:} \\ When CNNs Meet Random RNNs: Towards Multi-Level Analysis for RGB-D Object and Scene Recognition}

\author[1]{Ali \snm{Caglayan}\corref{cor1}} 
\cortext[cor1]{Corresponding author:} 
\ead{firstname.lastname@aist.go.jp}
\author[1]{Nevrez \snm{Imamoglu}}
\author[2]{Ahmet Burak \snm{Can}}
\author[1]{Ryosuke \snm{Nakamura}}

\address[1]{National Institute of Advanced Industrial Science and Technologhy (AIST), Tokyo, Japan}
\address[2]{Department of Computer Engineering, Hacettepe University, Ankara, Turkey}

\maketitle

This supplementary document provides implementational and experimental details of the proposed method. It first gives a detailed data preparation pipeline in Section~\ref{sec:dataPreparation}. Then, the schematic overview of backbone pretrained CNN models and their level-wise extraction points are presented in Section~\ref{sec:backboneModels}. Next, it reports comparative computation time and memory profiling of these models in Section~\ref{sec:profiling}. After that, effect of multi-level RNN structure over single-level RNN structure has been evaluated empirically in Section ~\ref{sec:multilevelRNNEffect}. Finally, empirical performance of different fusion strategies and perfomance of finetuned CNN-only semantic features without RNN have been presented in Section~\ref{sec:fusionStrategies} and Section~\ref{sec.cnnOnly}, respectively.

\section{Data Preparation} \label{sec:dataPreparation}
Following common practices for preprocessing, we resize RGB images to 256x256 dimensions according to bilinear transformation and apply center cropping to get 224x224 dimensional images. Then, we apply commonly used z-score standardization on the input data by using mean and standard-deviation of the ImageNet \citep{Deng_Imagenet_CVPR_2009}.

For depth domain, we first need appropriate RGB-like representation of depth data to leverage the power of pretrained CNN models over the large-scale RGB dataset of the ImageNet. To do so, there are several ways to represent depth data as RGB-like images such as HHA method of \cite{Gupta_ECCV_2014} (i.e. using horizontal and vertical observation values and angle of the normal to common surface), ColorJet work by \cite{Eitel_IROS_2015} (i.e. mapping depth values to different RGB color values), or commonly used surface normal based colorization as in \citep{Bo_IROS_2011, Caglayan_ECCVW_2018}. In this work, we prefer to use the colorization technique based on surface normals, as it confirms its effectiveness in our previous work \citep{Caglayan_ECCVW_2018}. However, unlike surface normal estimation from depth maps without camera parameters in \citep{Caglayan_ECCVW_2018}, we improve this in a more accurate way by estimating surface normals on 3D point clouds that have been computed using depth maps and camera intrinsic parameters. To address the issue of missing depth values, we first apply a fast vectorized depth interpolation by applying a median filter through a $5\times5$ neighborhood to reconstruct missing values in noisy depth inputs. Then, 3D point cloud estimation by using camera intrinsic constants and surface normal calculation on point clouds are followed, respectively. After this, the common approach is scaling surface normals to map values to the $0 - 255$ range to fit RGB image processing. However, since such an approach of mapping from floating point to integer values leads to a loss of information, we use these normal vectors as is without performing further quantization or scaling. Furthermore, unlike in RGB input processing, we apply resizing operation on these RGB-like depth data using the nearest neighborhood based interpolation rather than bilinear interpolation. Because the latter may lead to more distortion in geometric structure of a scene. Moreover, nearest neighbor interpolation is more suitable to the characteristics of depth data by providing a better separability between foreground and background in a scene. When applying z-score standardization to depth domain, we use the standard-deviation of the ImageNet as in RGB domain. However, we use zero-mean instead of the ImageNet mean as normal vectors are in the range of $[-1, 1]$ without the need for zero-mean shifting.
\begin{figure*}[!t]
	\centering
	\includegraphics[width=0.98\textwidth]{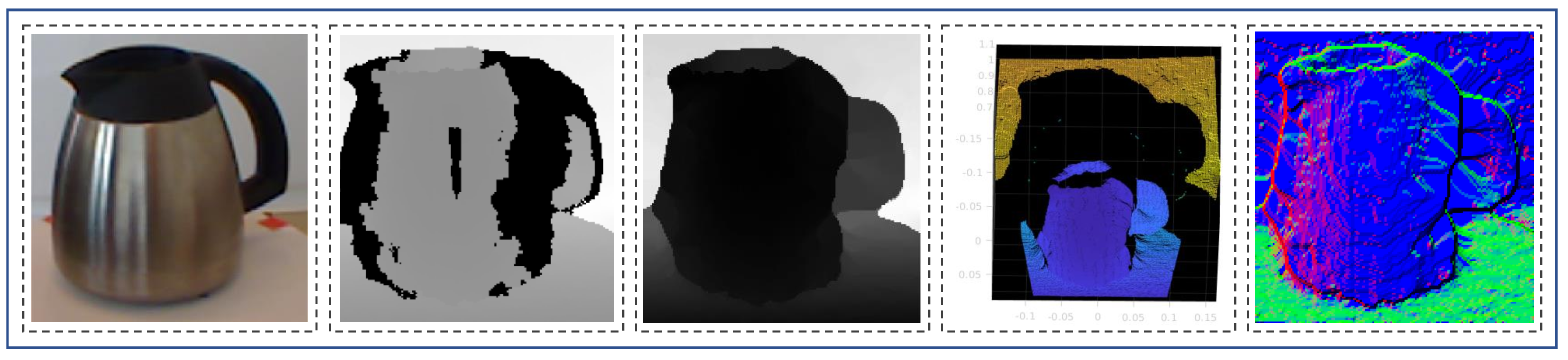}
	\caption{Illustration of data preparation pipeline. From right to left images are; RGB, depth map (contrast enhanced for visualization), depth map after interpolation process (contrast enhanced for visualization), 3D point cloud, and finally colorized depth based on surface normals to be given as input to the proposed model.}
	\label{fig:dataPreparation}
\end{figure*}
\section{Backbone Models}
\label{sec:backboneModels}
\begin{figure*}[!t]
	\centering
	\includegraphics[width=0.75\textwidth]{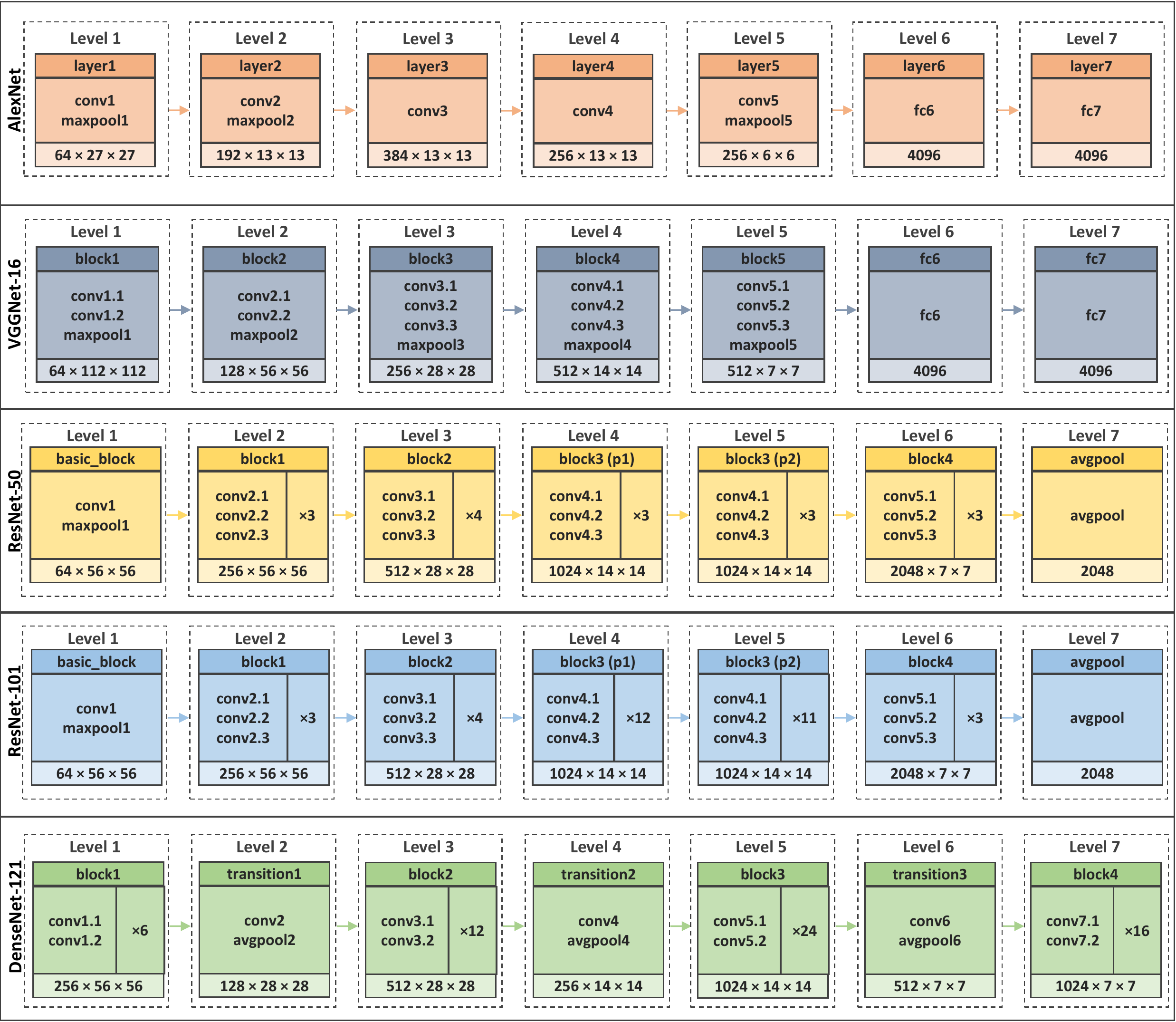}
	\caption{Schematic overview of CNN models and their level-wise extraction points based structures. Each level of schematic view shows name of the level, operations performed in the level with the number of them if exist (for ResNet \citep{He_CVPR_2016} and DenseNet \citep{Huang_CVPR_2017} models), and dimensions of the activation output.}
	\label{fig:Models}
\end{figure*}
In this work, we employ several available pretrained models of PyTorch including AlexNet \citep{Krizhevsky_NIPS_2012}, VGGNet \citep{Simonyan_ICLR_2015} (specifically VGGNet-16 model with batch normalization), ResNet \citep{He_CVPR_2016} (specifically ResNet-50 and ResNet-101 models), and DenseNet \citep{Huang_CVPR_2017}. We extract features from seven different levels of CNN models. The models investigated in this study with the feature extraction levels are shown in Fig. \ref{fig:Models}. Each level of the CNN models shows name of the level, operations performed in the level with the number of them if exist (for ResNet \citep{He_CVPR_2016} and DenseNet \citep{Huang_CVPR_2017} models), and dimensions of the activation output.

\section{Computation Time and Memory Profiling on Different Models}
\label{sec:profiling}
\begin{table*}[!hb]
	\caption{Average computational time and memory overhead for overall data processing and model learning on two splits of Washington RGB-D dataset. Results cover both of train and test phases together.}  
	\label{table:cnnProfiling}    
	\centering
	\setlength{\tabcolsep}{0.9em} 
	\def\arraystretch{1.2}
	\begin{adjustbox}{width=0.92\linewidth}
		\begin{tabular}{l|c|c|c|c|c|c|c|}
			\cline{2-8}
			& \multicolumn{3}{c|}{Time \textit{(hh:mm:ss)}}                                                                                                                                                                & \multicolumn{4}{c|}{Memory}                                                                                                                                                                                                  \\ \hline
			\multicolumn{1}{|l|}{\multirow{2}{*}{Model}} & \multirow{2}{*}{\begin{tabular}[c]{@{}c@{}}Feature Extraction\\ (CNN-RNN Stages)\end{tabular}} & \multirow{2}{*}{\begin{tabular}[c]{@{}c@{}}Classification\\ (SVMs)\end{tabular}} & \multirow{2}{*}{Overall} & \multirow{2}{*}{\begin{tabular}[c]{@{}c@{}}CNN-Stage\\ (GPU)\end{tabular}} & \multicolumn{2}{c|}{RNN-Stage (CPU)}             & \multicolumn{1}{l|}{\multirow{2}{*}{\begin{tabular}[c]{@{}l@{}}Overall\\ (CPU)\end{tabular}}} \\ \cline{6-7}
			\multicolumn{1}{|l|}{}                       &                                                                                                &                                                                                  &                          &                                                                            & \multicolumn{1}{c|}{Pool Weights} & RNN Weights & \multicolumn{1}{l|}{}                                                                         \\ \hline
			\multicolumn{1}{|l|}{AlexNet}                & 00:07:41                                                                                       & 00:28:33                                                                         & 00:36:14                 & 1115 MB                                                                    & \multicolumn{1}{c|}{772.1 kB}     & 4.2 GB      & 12.6 GB                                                                                       \\ \hline
			\multicolumn{1}{|l|}{VGGNet-16}              & 00:21:21                                                                                       & 00:36:42                                                                         & 00:58:03                 & 9259 MB                                                                    & \multicolumn{1}{c|}{8.6 MB}       & 4.8 GB      & 11.8 GB                                                                                       \\ \hline
			\multicolumn{1}{|l|}{ResNet-50}              & 00:16:23                                                                                       & 00:38:36                                                                         & 00:54:59                 & 6067 MB                                                                    & \multicolumn{1}{c|}{9.6 MB}       & 5.1 GB      & 10.8 GB                                                                                       \\ \hline
			\multicolumn{1}{|l|}{ResNet-101}             & 00:19:08                                                                                       & 00:40:33                                                                         & 00:59:41                 & 8795 MB                                                                    & \multicolumn{1}{c|}{9.6 MB}       & 5.1 GB      & 11.8 GB                                                                                       \\ \hline
			\multicolumn{1}{|l|}{DenseNet-121}           & 00:17:02                                                                                       & 00:26:47                                                                         & 00:43:49                 & 8821 MB                                                                    & \multicolumn{1}{c|}{8.3 MB}       & 5.4 GB      & 13.3 GB                                                                                       \\ \hline
		\end{tabular}
	\end{adjustbox}
\end{table*}
We evaluate different baseline CNN models within our framework in terms of computational time and memory requirements. We evaluate the proposed framework in two parts: (\textit{i}) Feature extraction containing CNN-RNN stages and (\textit{ii}) Classification where a model based on the extracted features is learnt to distinguish the different classes. The batch size is set to 64 for all the models. Table \ref{table:cnnProfiling} reports computational times and memory workspaces for the whole data processing ($41,877$ images) on Washington RGB-D dataset. The results here are the average results of two splits on RGB images. There is additional cost for depth data processing as it is required to colorize them. The results on this table cover the overall processing and classification of all $7$ level features. Moreover, it should be noted that classification time covers both training and testing processes, in which training takes the main computational burden. Therefore, the main cost in terms of processing time comes from training SVM models that works on CPU for $7$ times. The process for only a single optimum level would reduce the computational time to a ratio of seven approximately. Hence, using a single optimum level or fusion of selected levels can be efficient enough in terms of time and memory requirements while presenting sufficient representations.

\section{Effect of Multi-Level RNN Structure}
\label{sec:multilevelRNNEffect}
\begin{figure}[!t]
	\centering
	\includegraphics[width=0.6\columnwidth, keepaspectratio]{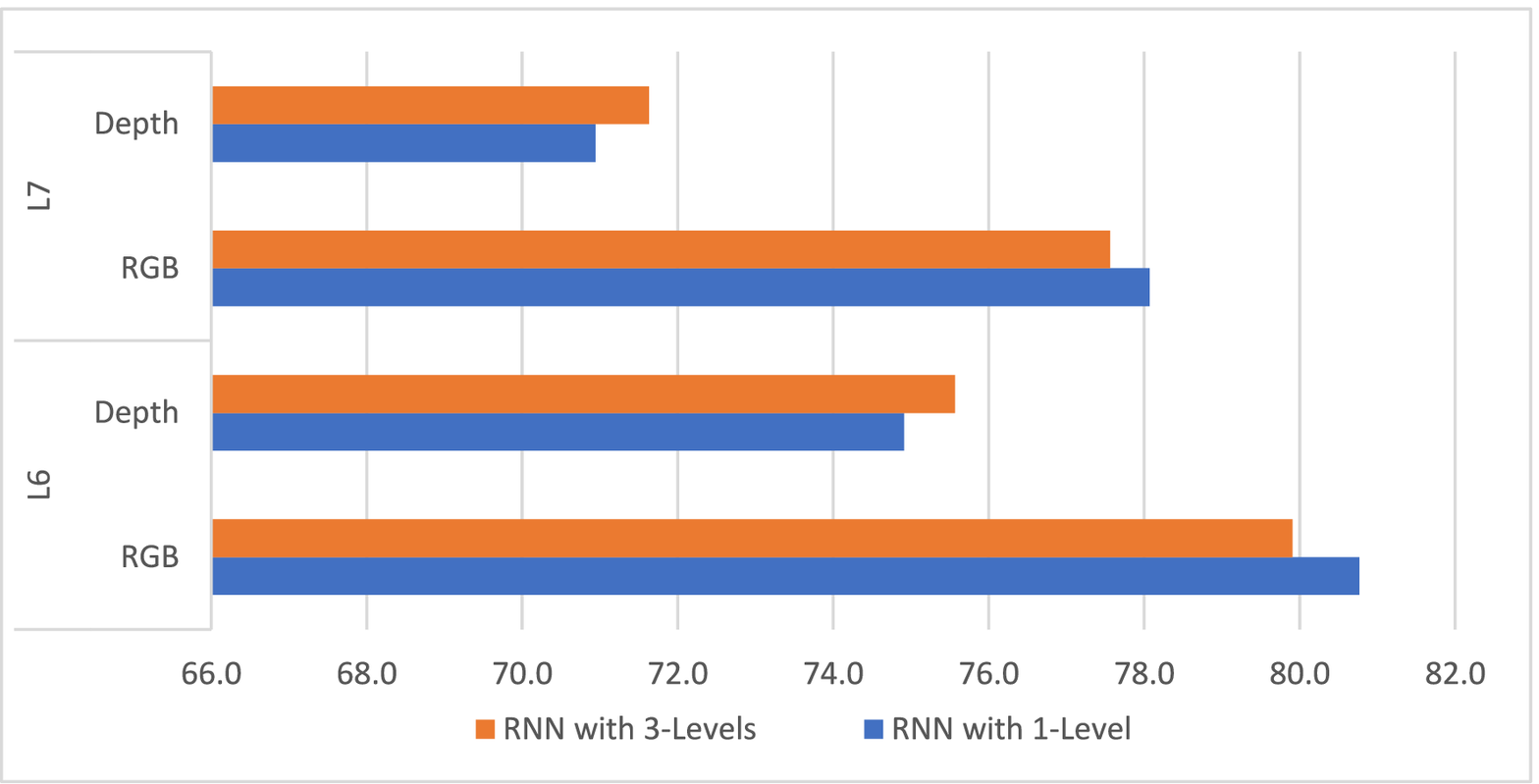}
	\caption{Comparison of single-level and multi-level RNNs on two different CNN activations (L6 and L7) of AlexNet. The horizontal axis shows average accuracy performances (\%) on two splits of Washington RGB-D dataset.}
	\label{fig:multilevelRNN}
\end{figure}

An RNN in this study is of one-level structure with a single parent computation, which is obviously computationally fast comparing to the multi-level structural RNNs. Furthermore, in this way, it provides an ease of use with no need of further processing for fixing the required input forms. However, in order to testify the performance of single-level RNNs over multiple-level RNNs, we analyze the comparative accuracy performances of 1-level RNNs together with 3-levels RNNs (see Fig. 2 in main paper). To this end, we conduct experiments on two CNN activation levels with highest semantic information (L6 and L7) of the baseline model of AlexNet. The average results of two splits for both of RGB and depth data are shown in Fig. \ref{fig:multilevelRNN}. The results show that RNN with 1-level performs better than RNN with 3-levels on RGB data, while 3-levels of RNN is better than 1-level of RNN on depth data. The better performance of RNN with 3-levels on depth data might be due to the use of a pretrained CNN model based on the RGB data of ImageNet. Hence, further processing might provide more representative information for depth data in that way. Therefore, this difference might be diminished or turn in favor of 1-level RNNs in the use of finetuned CNNs for depth modality as well. Overall, considering both RGB and depth data together, RNNs with 1-level are better in terms of accuracy performance as well.
\begin{table*}[!t]
	\caption{Average accuracy performance of different fusion combinations on the best four levels using Washington RGB-D dataset (\%).}  
	\label{table:levelFusions}    
	\centering
	\setlength{\tabcolsep}{0.9em} 
	\def\arraystretch{1.2}
	\begin{adjustbox}{width=0.9\linewidth}
		\begin{tabular}{cl|ll|ll|ll|}
			\cline{3-8}
														  &                       	& \multicolumn{2}{c|}{AlexNet}                         & \multicolumn{2}{c|}{DenseNet-121}                    & \multicolumn{2}{c|}{ResNet-101}                      \\ \cline{3-8} 
														  &                       	& \multicolumn{1}{c}{RGB} & \multicolumn{1}{c|}{Depth} & \multicolumn{1}{c}{RGB} & \multicolumn{1}{c|}{Depth} & \multicolumn{1}{c}{RGB} & \multicolumn{1}{c|}{Depth} \\ \hline
			\multicolumn{1}{|c|}{\multirow{4}{*}{Single}} & LB1                   	& 81.4 $\pm\hfil$ 1.8              & 83.5 $\pm\hfil$ 2.2                 & 91.0 $\pm\hfil$ 1.2              & 86.8 $\pm\hfil$ 2.1                 & 91.1 $\pm\hfil$ 1.0              & 87.1 $\pm\hfil$ 2.7                 \\
			\multicolumn{1}{|c|}{}                        & LB2                   	& 81.1 $\pm\hfil$ 2.1              & 83.3 $\pm\hfil$ 2.2                 & 89.7 $\pm\hfil$ 1.0              & 86.2 $\pm\hfil$ 2.3                 & 90.5 $\pm\hfil$ 1.6              & 86.9 $\pm\hfil$ 2.6                 \\
			\multicolumn{1}{|c|}{}                        & LB3                   	& 79.2 $\pm\hfil$ 2.4              & 83.2 $\pm\hfil$ 2.3                 & 89.5 $\pm\hfil$ 1.5              & 85.0 $\pm\hfil$ 2.1                 & 89.2 $\pm\hfil$ 1.3              & 85.5 $\pm\hfil$ 2.2                 \\
			\multicolumn{1}{|l|}{}                        & LB4                   	& 78.8 $\pm\hfil$ 2.5              & 82.6 $\pm\hfil$ 2.3                 & 85.7 $\pm\hfil$ 1.8              & 82.5 $\pm\hfil$ 2.0                 & 88.5 $\pm\hfil$ 2.2              & 83.8 $\pm\hfil$ 2.1                 \\ \hline
			\multicolumn{1}{|c|}{\multirow{3}{*}{Concats}}& LB1 + LB2             	& \textbf{83.0 $\pm\hfil$ 1.9}     & 84.0 $\pm\hfil$ 2.4                 & 90.4 $\pm\hfil$ 1.0              & 86.8 $\pm\hfil$ 2.1                 & 91.1 $\pm\hfil$ 1.5              & 87.0 $\pm\hfil$ 2.7                 \\
			\multicolumn{1}{|l|}{}                        & LB1 + LB2 + LB3       	& 82.5 $\pm\hfil$ 2.0              & 83.8 $\pm\hfil$ 2.3                 & 90.0 $\pm\hfil$ 1.5              & \textbf{86.9 $\pm\hfil$ 2.1}        & 91.5 $\pm\hfil$ 1.3              & 87.0 $\pm\hfil$ 2.7                 \\
			\multicolumn{1}{|l|}{}                        & LB1 + LB2 + LB3 + LB4 	& 82.7 $\pm\hfil$ 2.0              & 84.2 $\pm\hfil$ 2.4                 & 90.4 $\pm\hfil$ 1.3              & 86.8 $\pm\hfil$ 2.3                 & 90.7 $\pm\hfil$ 1.7              & 87.0 $\pm\hfil$ 2.7                 \\ \hline
			\multicolumn{1}{|c|}{\multirow{3}{*}{SVM Avg Voting}}& LB1 + LB2        & 82.8 $\pm\hfil$ 1.9              & 84.1 $\pm\hfil$ 2.3                 & 91.2 $\pm\hfil$ 1.0              & 86.8 $\pm\hfil$ 2.2                 & 92.2 $\pm\hfil$ 1.0              & 87.0 $\pm\hfil$ 2.7                 \\
			\multicolumn{1}{|l|}{}                        & LB1 + LB2 + LB3       	& 82.7 $\pm\hfil$ 2.1              & 84.0 $\pm\hfil$ 2.4                 & \textbf{91.5 $\pm\hfil$ 1.1}     & 86.7 $\pm\hfil$ 2.2                 & \textbf{92.3 $\pm\hfil$ 1.0}     & \textbf{87.2 $\pm\hfil$ 2.5}        \\
			\multicolumn{1}{|l|}{}                        & LB1 + LB2 + LB3 + LB4 	& 82.9 $\pm\hfil$ 2.1              & \textbf{84.6 $\pm\hfil$ 2.4}        & 91.2 $\pm\hfil$ 1.0              & 86.5 $\pm\hfil$ 2.2                 & 92.0 $\pm\hfil$ 1.2              & 87.1 $\pm\hfil$ 2.5                 \\ \hline
			\multicolumn{8}{l}{LB1, LB2, LB3, LB4: Best, Second best, Third best, and Fourth best performing level}
		\end{tabular}
	\end{adjustbox}
\end{table*}
\section{Empirical Performance of Different Fusion Strategies}
\label{sec:fusionStrategies}
We have shown that a fixed pretrained CNN model together with random RNN already achieves impressive results on a single level. Likewise, when such pretrained models are finetuned on depth data, the results are boosted greatly. The best single levels for RGB and depth data, respectively, are L4 and L5 for AlexNet; L5 and L6 for ResNet-101; and  L6 and L7 for DenseNet-121. Next, to further improve accuracy performances, we investigate empirical accuracy analysis of multi-level fusions using fixed pretrained CNN models on RGB data and finetuned CNN models on depth data. In this work, in addition to the feature concatenation as in our previous work \citep{Caglayan_ECCVW_2018}, we also apply average voting based on SVM confidence scores on the best performing levels. Table \ref{table:levelFusions} reports the average accuracy on the all 10 train/test splits of Washington RGB-D dataset for AlexNet, DenseNet-121, and ResNet-101. The table shows the top four level results (best levels) for each modality and their fusion combinations. The best four levels (LB1, LB2, LB3, LB4) of AlexNet are (L4, L5, L6, L3) on RGB data and (L5, L6, L7, L4) on depth data, respectively; of ResNet-101 are (L5, L6, L4, L7) on RGB data and (L6, L7, L5, L4) on depth data, respectively; and for DenseNet-121 these levels are (L6, L5, L7, L4) on RGB and (L7, L6, L5, L4) on depth data, respectively. As can be seen from the table, a single level has already produced very good results. Since both model structures and data modality characteristics are different, the best results for each column generally vary depending on the data type and the used model. Nevertheless, in general, average voting on SVM confidence scores gives better results comparing to feature concatenation. We can also see that fusion of more levels does not necessarily increase the accuracy success. In general, the optimum results are achieved with SVM average voting of the best three levels for ResNet-101 and DenseNet-121 models. 

\section{Performance of Finetuned CNN-only Semantic Features} \label{sec.cnnOnly}
In our previous work \citep{Caglayan_ECCVW_2018}, we analyze the comparison of CNN-only features with the CNN-RNN incorporation-based features on best performing middle layers and show the advantages of using CNN-RNN. On the other hand, as we stated in the main paper, features obtained from semantically rich information-based final layers is often utilized in many approaches \citep{Razavian_CVPRW_2014, Schwarz_ICRA_2015, Girshick_CVPR_2014, Sermanet_ICLR_2014, Farabet_TPAMI_2013}. Therefore, here we also extract features from the final semantic layer, level 7, and classify them with a linear SVM and report the results in Table~\ref{table:SemanticCNNs}. We use the finetuned models of AlexNet, DenseNet-121, and ResNet-101 models. RGB-D results are computed using the average RGB and depth SVM confidence scores. The results confirm that the backbone models produce impressive performance by finetuning the CNN models properly (see finetuning parameter setups in the main paper for more details). In addition, we can see that in case of CNN features powered by the incorporation of RNNs with the stronger representations (see object recognition performance in the main paper), the results are better compared to the CNN-only semantic features. Moreover, there is no any extra training cost, since RNNs in our model do not require training.
\begin{table}[!h]
	\caption{Average accuracy performance of semantic CNN features (L7 features) without RNN using different finetuned baseline models on Washington RGB-D dataset (\%).}
	\begin{center}
		\setlength{\tabcolsep}{0.9em} 
		\def\arraystretch{1.2}
		\begin{adjustbox}{width=0.4\columnwidth}
			\begin{tabular}{ lccc }
				\hline
				Accuracy			& RGB 								& Depth 							& RGB-D \\ \hline \hline
				AlexNet 			& 74.3 $\pm\hfil$ 2.8 				& 83.1 $\pm\hfil$ 2.2 	 			& 89.0 $\pm\hfil$ 1.9 		\\ 
				DenseNet-121    		& 88.8 $\pm\hfil$ 1.2 				& 86.7 $\pm\hfil$ 2.1 	 			& 93.2 $\pm\hfil$ 1.6 		\\ 
				ResNet-101   		& 89.0 $\pm\hfil$ 0.8 				& 86.7 $\pm\hfil$ 2.5 	 			& 92.8 $\pm\hfil$ 1.4 		\\ 
				\hline
			\end{tabular}
		\end{adjustbox}
		\label{table:SemanticCNNs}
	\end{center}
\end{table}


%





\bibliographystyle{model2-names}
\bibliography{cviu}